\documentclass[10pt,twocolumn,letterpaper]{article}

\usepackage{iccv}
\usepackage{graphicx}
\usepackage{times}
\usepackage{epsfig}
\usepackage{algorithm}
\usepackage{multirow}
\usepackage{mathtools,xparse}
\usepackage{siunitx}
\usepackage{stmaryrd}
\usepackage{amsmath}
\usepackage[noend]{algpseudocode}
\usepackage{placeins}
\usepackage{mdframed}
\usepackage{amsmath,amssymb} % define this before the line numbering.
\usepackage{color}
\usepackage{subcaption}
\usepackage{amsthm}
\usepackage[table]{xcolor}

\theoremstyle{definition}
\newtheorem{defn}{Definition}[section]
\newcommand*\rot{\rotatebox{90}}

\makeatletter
\def\BState{\State\hskip-\ALG@thistlm}
\makeatother

\algnewcommand\algorithmicforeach{\textbf{for each}}
\algdef{S}[FOR]{ForEach}[1]{\algorithmicforeach\ #1\ \algorithmicdo}

\newcommand*\trans{\text{T}}
\newcommand*\Instances{h}
\newcommand*\CompoundModel{h_{\cup}}
\newcommand*\Points{\mathcal{P}}

\newcommand*\Point{\mathbf{p}}

\newcommand*\DistanceFunction{\phi}

\usepackage[pagebackref=true,breaklinks=true,letterpaper=true,colorlinks,bookmarks=false]{hyperref}

\iccvfinalcopy % *** Uncomment this line for the final submission
 
\def\iccvPaperID{5018} % *** Enter the CVPR Paper ID here

\ificcvfinal\fi
\begin{document}

%%%%%%%%% TITLE
\title{Progressive-X: Efficient, Anytime, Multi-Model Fitting Algorithm} 

\author{Daniel Barath$^{12}$, and Jiri Matas$^{1}$\\
$^1$ Centre for Machine Perception, Department of Cybernetics \\
  Czech Technical University, Prague, Czech Republic \\
  $^2$ Machine Perception Research Laboratory, 
  MTA SZTAKI, Budapest, Hungary \\
    {\tt\small barath.daniel@sztaki.mta.hu}
}

\maketitle
%\thispagestyle{empty}

%%%%%%%%% ABSTRACT
\begin{abstract}
The Progressive-X algorithm, Prog-X in short, is proposed for geometric multi-model fitting. 
The method interleaves sampling and consolidation of the current data interpretation via repetitive hypothesis proposal, fast rejection, and integration of the new hypothesis into the kept instance set by labeling energy minimization. 
Due to exploring the data progressively, the method has several beneficial properties compared with the state-of-the-art. 
First, a clear criterion, adopted from RANSAC, controls the termination and stops the algorithm when the probability of finding a new model with a reasonable number of inliers falls below a threshold. 
Second, Prog-X is an any-time algorithm. Thus, whenever is interrupted, e.g.\ due to a time limit, the returned instances cover real and, likely, the most dominant ones.
The method is superior to the state-of-the-art in terms of accuracy in both synthetic experiments and on publicly available real-world datasets for homography, two-view motion, and motion segmentation. 
% Also, we detected and manually assigned correspondences to homographies in every image pair of the Strecha dataset for fitting multiple homographies. The manual annotation will be made available with the publication.  
\end{abstract}

\section{Introduction} 

The multi-class multi-model fitting problem is to interpret a set of input points as the mixture of noisy observations originating from multiple model instances which are not necessarily of the same class. 
Examples of this problem are the estimation of $k$ planes and $l$ spheres in a 3D point cloud; multiple line segments and circles in 2D edge maps; a number of homographies or fundamental matrices from point correspondences; or multiple motions in videos. 
Robustness is achieved by considering the assignment to one or multiple outlier classes.

\begin{figure}
    \centering
    %\begin{subfigure}[t]{0.99\columnwidth}
   	% 	\centering
    %    \includegraphics[width=\textwidth]{assets/Multi-model_fitting_approaches.png}
    %    \caption{The multi-model fitting pipeline used in the most recent papers.
    %    First, a number of model instances are generated. Finally, an optimization procedure selects the ones which interpret the input data the most. }
    %    \label{fig:approach_traditional}
    %\end{subfigure}\\
    %
    %\begin{subfigure}[t]{0.99\columnwidth}
    	\centering
        \includegraphics[width=0.99\columnwidth]{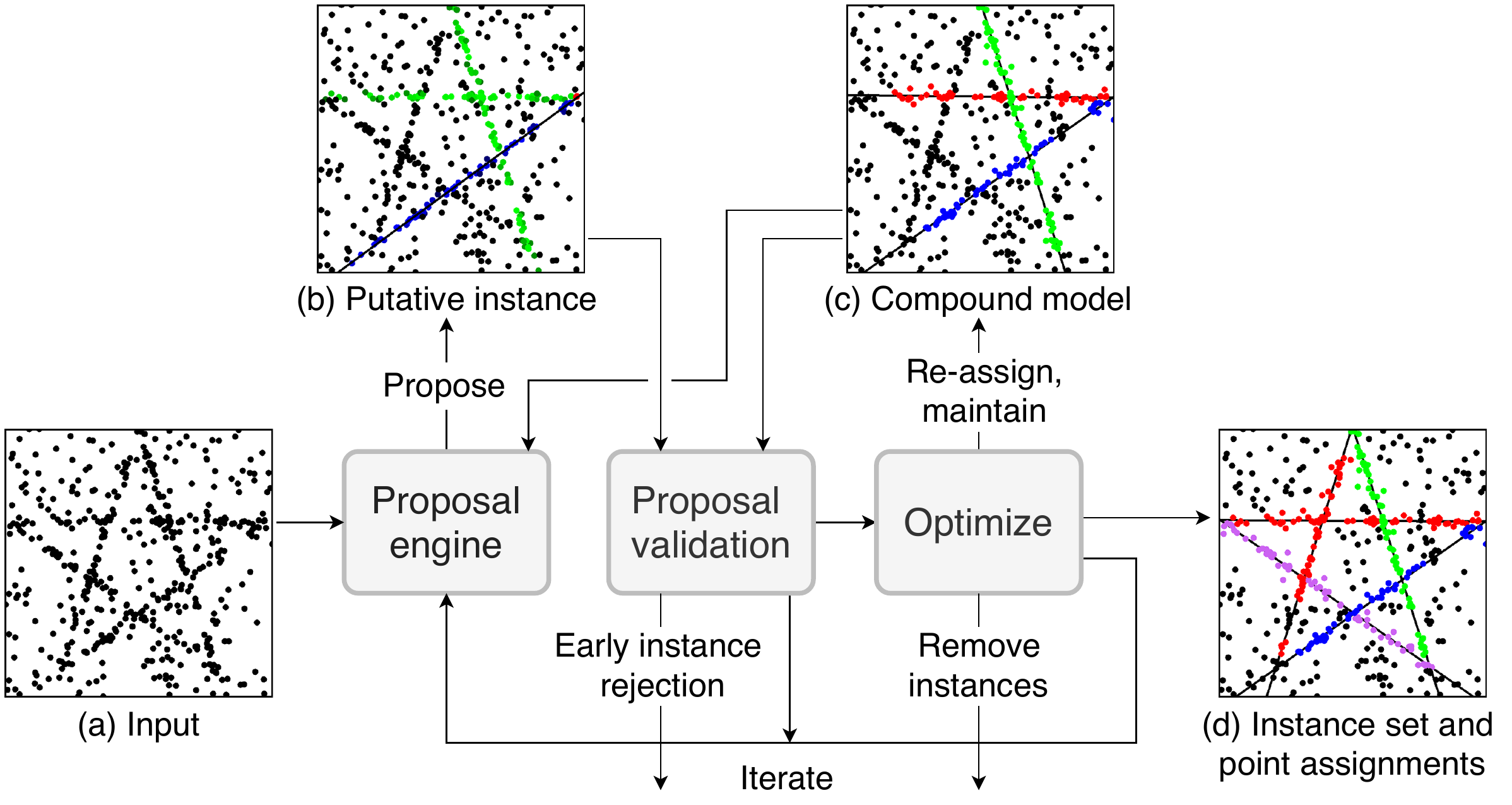}
        \caption{The Prog-X multi-model fitting pipeline gets a point set as input (a). 
        It then repeats three steps.
        \textit{First}, proposal of putative hypotheses (b) considering the active instance set (c), i.e.\ the compound model.
        In (b), the blue points are inliers of the putative instance, and the green ones are that of the compound one.
        \textit{Second}, fast rejection of redundant hypotheses. 
        \textit{Third}, 
        optimization by integrating new hypotheses, re-assigning points,
        maintaining model parameters, and
        removing unnecessary instances.
        In (c) and (d), the color codes the assignment to instances. 
        %\textit{First}, it proposes a putative instance, blue point on (a).
        %In each iteration, first, it generates an instance which has the most points not shared (blue points) with the previous proposals (green points). 
        %Finally, an optimization is applied to all instances assigning the points to models.
        }
        \label{fig:approach_new}
    %\end{subfigure}
    %\caption{The traditional and proposed multi-model fitting pipelines.}
\end{figure}

Multi-model fitting has been studied since the early sixties. 
The Hough-transform~\cite{vc1962method,illingworth1988survey} perhaps is the first popular method for finding multiple instances of a single class~\cite{guil1997lower,matas2000robust,rosin1993ellipse,xu1990new}. 
The RANSAC~\cite{fischler1981random} algorithm was extended as well  to deal with multiple instances. 
Sequential RANSAC~\cite{vincent2001detecting,kanazawa2004detection} detects instances one after another by searching in the point set from which the inliers of the detected instances have been removed.  
The greedy approach makes RANSAC a powerful tool for finding a single instance, is a drawback in multi-instance setting.  Points are assigned not to the best but to the first instance, typically the one with the largest support, for which they cannot be deemed outliers. 
MultiRANSAC~\cite{zuliani2005multiransac} forms compound hypotheses about $n$ instances.
Besides requiring the number $n$ of the instances to be known a priori, 
the processing time is high,  since samples of size $n$ times $m$ in are drawn in each iteration, where $m$ is the number of points required for estimating a model instance and finding an all-inlier simple is rare.
Nevertheless, RANSAC-based approaches have a desirable property of interleaving the hypothesis generation and verification steps. Moreover, they have a justifiable termination criterion based on the inlier-outlier ratio in the data which provides a probabilistic guarantee of finding the best instance. 

\begin{figure*}
    \centering
    \begin{subfigure}[t]{0.99\columnwidth}
   	 	\centering
        \includegraphics[width=0.49\columnwidth]{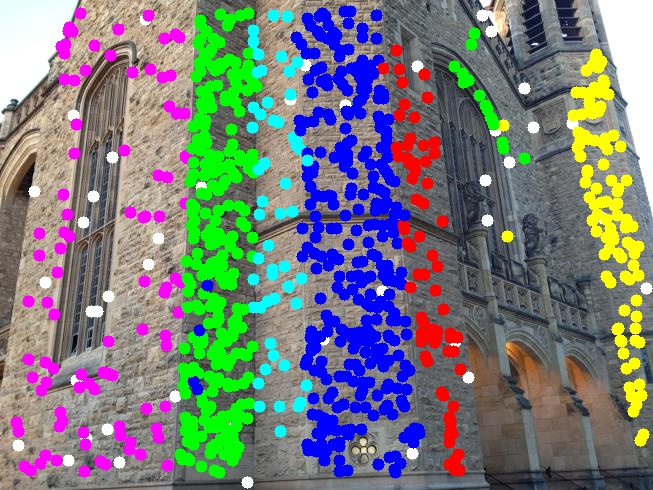}
        \includegraphics[width=0.49\columnwidth]{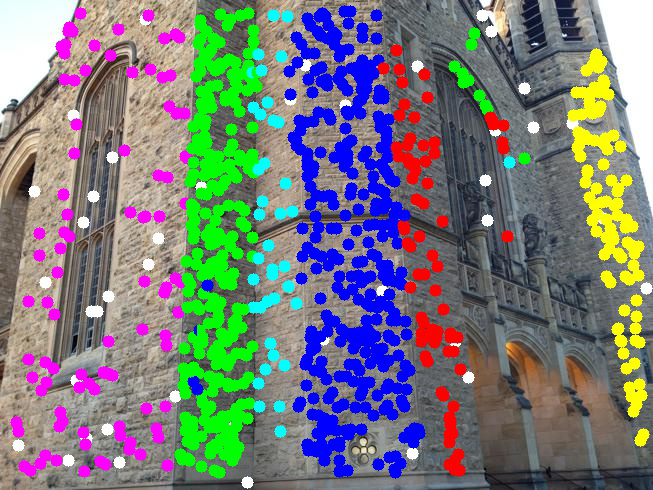}
        \caption{\textit{homographies}; (left) best, 4.8\%; (right) worst, 5.6\%}
        \label{fig:real_results_H_1}
    \end{subfigure}
    \begin{subfigure}[t]{0.99\columnwidth}
   	 	\centering
        \includegraphics[width=0.49\columnwidth]{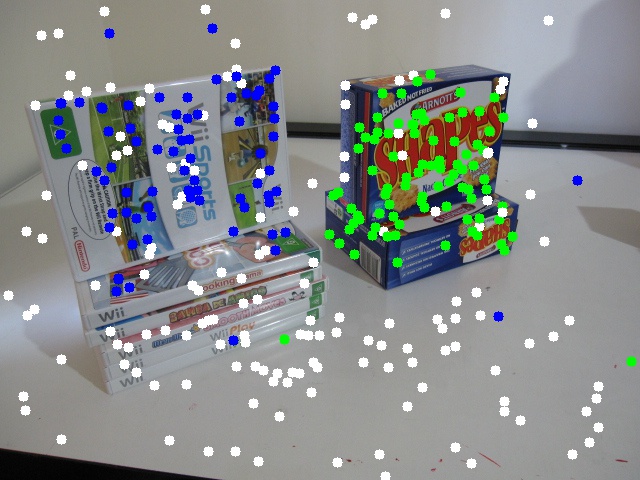}
        \includegraphics[width=0.49\columnwidth]{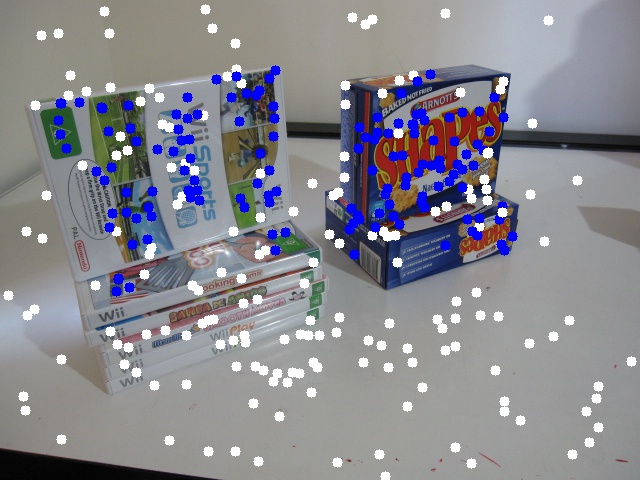}
        \caption{\textit{two-view motions}; (left) best, 2.4\%; (right) worst, 27.4\%}
        \label{fig:real_results_F}
    \end{subfigure}\\
    \begin{subfigure}[t]{0.99\columnwidth}
   	 	\centering
        \includegraphics[width=0.49\columnwidth]{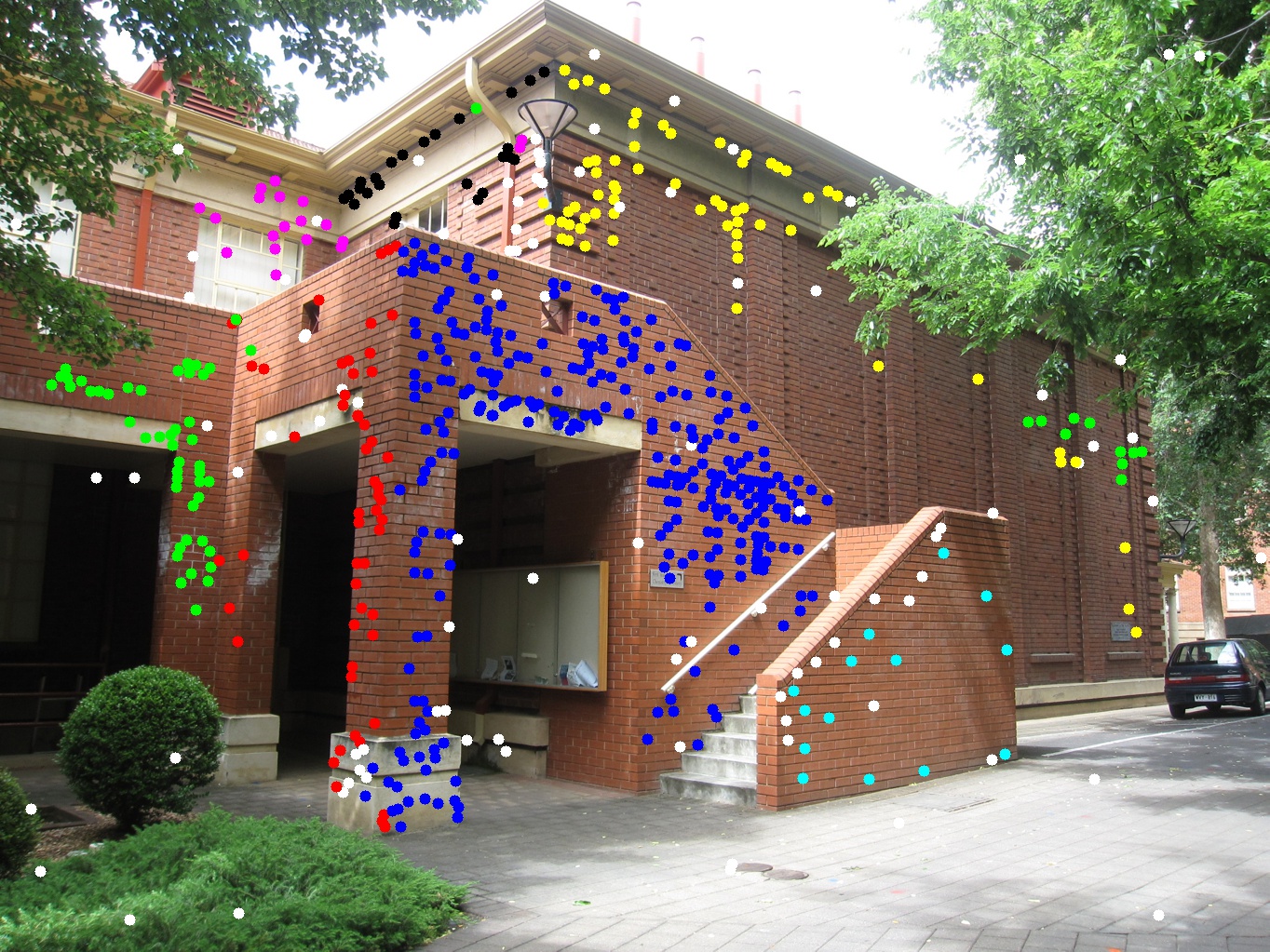}
        \includegraphics[width=0.49\columnwidth]{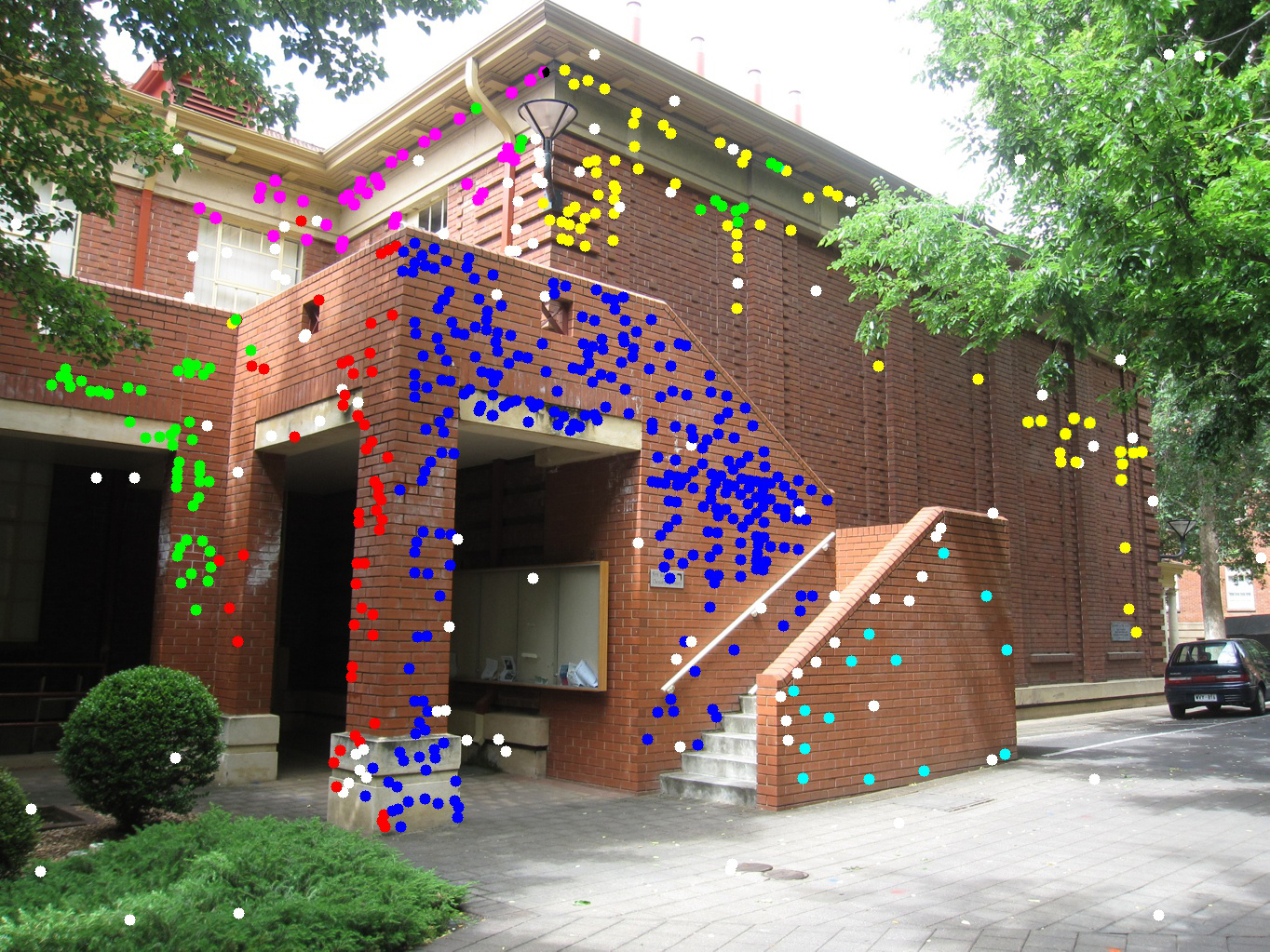}
        \caption{\textit{homographies}; (left) best, 5.9\%; (right) worst, 12.6\%}
        \label{fig:real_results_H_2}
    \end{subfigure}
    \begin{subfigure}[t]{0.99\columnwidth}
   	 	\centering
        \includegraphics[width=0.49\columnwidth]{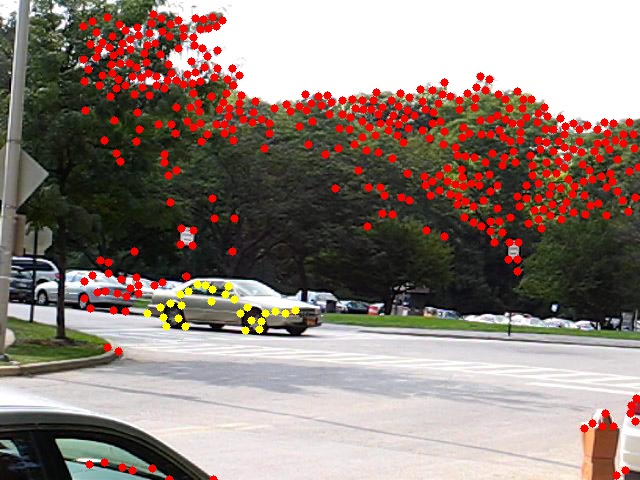}
        \includegraphics[width=0.49\columnwidth]{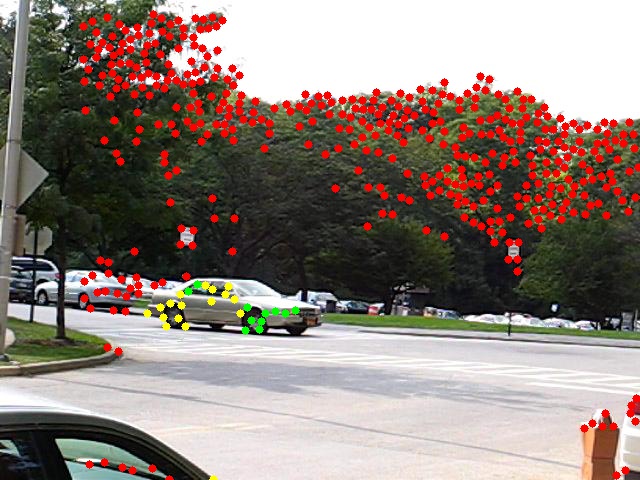}
        \caption{\textit{motions}; (left) best, 0.0\%; (right) worst, 2.4\%}
        \label{fig:real_results_m}
    \end{subfigure}
    \caption{Prog-X -- quality and stability of the results on selected problems: homography estimation (a,c), multiple epipolar geometries (b) and rigid motion (d). The best (left) and worst (right) results for five runs with fixed parameters, overlayed on one of the images. White points have been labeled as outliers, color codes the assignment to an instance. % TODO: make images with the same colors
    Misclassification errors, i.e.\ the fraction of points assigned to the wrong cluster, are shown as percentages. 
    Results in Table~\ref{tab:summary_table} show that competing algorithms are much less stable.
    }
    \label{fig:real_results}
\end{figure*}

Most recent approaches for multi-model fitting~\cite{wongiccv2011,isack2012energy,pham2014interacting,magri2014t,magri2015robust,wang2015mode,magri2016multiple,barath2018multix,amayo2018geometric} follow a two-step procedure, first, generating a number of instances using RANSAC-like hypothesis generation. Second, a subset of the generated hypotheses is selected interpreting the input data points the most.
This selection is done in various ways. 
For instance, a popular group of methods~\cite{isack2012energy,pham2014interacting,barath2018multix,amayo2018geometric} optimizes point-to-instance assignments by energy minimization using graph labeling techniques~\cite{boykov2004experimental}. 
The energy originates from point-to-instance residuals, label costs~\cite{delong2012minimizing}, geometric priors interacting among the models~\cite{pham2014interacting}, and from the spatial coherence of the points. %by considering the spatial relationships of the points.
Another group of methods uses preference analysis based on the distribution of the residuals of data points.~\cite{zhang2007nonparametric,magri2014t,magri2015robust,magri2016multiple}.
Also, there are techniques~\cite{wang2015mode,wang2018searching} approaching the problem as a hyper-graph partitioning where the instances are vertices, and the points are hyper-edges.

\textit{The common part of these algorithms} is the initialization step when model \textit{instances are generated blindly}, having no information about the data. 
As a consequence, it must be decided by the user whether to consider the worst case scenario and, thus, generate an unnecessarily high number of instances; or to use some rule of thumb, e.g.\ to generate twice the point number hypotheses. In practice, this is what is usually done. It, however, offers no guarantees for covering all the desired instances. 
The first method recognizing this problem was Multi-X~\cite{barath2018multix}. It added a new step to the optimization procedure to reduce the number of instances by replacing sets of labels by the corresponding density modes in the model parameter domain. 
Even though this step allows the generation of more initial instances than before without high computational demand, it still does not provide guarantees, especially, in case of low inlier ratio. 

The main contribution of the proposed Prog-X is its any-time nature which originates from interleaving sampling and consolidation of the current data interpretation. 
%due to alleviating the aforementioned problem by interleaving sampling and consolidation of the current data interpretation. 
This is done via repeated instance proposal minding the already proposed ones; fast rejection of redundant hypotheses; and integration of the new hypothesis into the kept instance set by energy minimization (see Fig.~\ref{fig:approach_new} for the basic concept).
Moreover, Prog-X adopts the probabilistic guaranties of RANSAC by progressively exploring the data.  
% and, due to progressively exploring the data, \textit{it is an any-time algorithm}. Therefore, whenever it is stopped, the returned hypotheses cover real and, likely, the most dominant instances.
We use a general instance-to-instance metric based on set overlaps which can be efficiently estimated by the min-hash method~\cite{broder1997resemblance} and modify the model quality function of RANSAC considering the existence of multiple model instances.
The method is tested both in synthetic experiments and on publicly available real-world datasets. 
It is superior to the state-of-the-art in terms of accuracy for homography, two-view motion, motion clustering, and simultaneous plane and cylinder fitting.

%\begin{verbatim}
%- PEARL and Multi-X are fine, but they suffer from the following weakness: the initial step, generating RANSAC-like initial hypothesis for the model instance, is done blindly, with no information about the structure of the datas. As a results the mehtods most cover the worst case, generating a large number of instances, even if only a few are needed
%- Contrast this with RANSAC, that samples only until it is unlikely that it can improve the so-far-the-best model, terminating in a data dependent manner.
%- We propose Progressive-X that alleviates the above-mentioned problem by interleaving sampling, and consolidation of the current data interpretation via fast reject of redundant new hypothesis or intergration of the new hypothesis in the model by energy minimization.
%- We get best of both worlds.
%Contributions:
%- anytime property
%- clear termination criterion
%Improve upon Multi-X by:
%- the integration of a new instance replace the mode-seeking 
%- use a general instnace-to-instance metric based on set overlap, efficiently 
%  estmated by minHash.
%- state of the art results
%\end{verbatim}

\section{Terminology and notation}

In this section, we discuss the most important concepts used in this paper. 
For the sake of generality, we consider multi-class multi-model fitting, thus, aiming to find multiple model instances not necessarily of the same model class. 
We adopt the notation from~\cite{barath2018multix}. 

\noindent
\textbf{Multi-class multi-instance model fitting.}
First, we show the problem through examples.  
A simple problem is finding a pair of \textit{line instances} $h_1, h_2 \in \mathcal{H}_l$ interpreting a set of 2D points $\Points \subseteq \mathbb{R}^2$.
Line class $\mathcal{H}_l$ is the space of lines $\mathcal{H}_l = \{(\theta_l, \phi_l, \tau_l), \theta_l = [\alpha \quad c]^\trans \}$ equipped with a distance function $\phi_l(\theta_l, p) = |\cos(\alpha) x + \sin(\alpha) y + c|\;$ ($p = [x \quad y]^\trans \in \Points$) and a function $\tau_l(p_1, ..., p_{m_l}) = \theta_l$ for estimating $\theta_l$ from $m_l \in \mathbb{N}$ data points. 
%Another example is the fitting of $n$ \textit{circle instances} $h_1, h_2, \cdots, h_n \in \mathcal{H}_c$ to 2D points. 
%The circle class $\mathcal{H}_c = \{(\theta_c, \phi_c, \tau_c), \theta_c = [c_x \quad c_y \quad r]^\trans\}$ is the space of circles, $\phi_c(\theta_c, p) = |r - \sqrt{(c_x - x)^2 + (c_y - y)^2}|$ is a distance function and $\tau_c(p_1, ..., p_{m_c}) = \theta_c$ is an estimator. 
%
\textit{Multi-line fitting} is the problem of finding multiple line instances $\{h_1, h_2, ...\} \subseteq \mathcal{H}_l$, while the \textit{multi-class} case is extracting a subset $\mathcal{H} \subseteq \mathcal{H}_\forall$, where $\mathcal{H}_\forall = \mathcal{H}_l \cup \mathcal{H}_c \cup \mathcal{H}_. \cup \cdots$. 
The set $\mathcal{H}_\forall$ is the space of all classes including that of lines and circles. 
Also, the formulation includes the outlier class $\mathcal{H}_o = \{(\theta_o, \phi_o, \tau_o), \theta_o = \varnothing \}$ where each instance has a constant distance to all points $\phi_o(\theta_o, p) = k$, $k~\in~\mathbb{R}^+$ and $\tau_o() = \varnothing$. 
% Note that considering multiple outlier classes allows the interpretation of outliers originating from different sources.  
%
%The \textit{multi-class model} is a space $\mathcal{H}_\forall = \bigcup \mathcal{H}_i$, where $\mathcal{H}_i = \{ (\theta_i, \phi_i, \tau_i) \; | \; d_i \in \mathbb{N}, \theta_i \in \mathbb{R}^{d_i}, \phi_i \in \Points \times \mathbb{R}^{d_i} \rightarrow \mathbb{R}, \tau_i : \PointsDesc \to \mathbb{R}^{d_i} \}$ is a single class, $\Points$ is the set of data points, $d_i$ is the dimension of parameter vector $\theta_i$, $\phi_i$ is the distance function and $\tau_i$ is the estimator of the $i$th class. 
%
The \textit{objective of multi-instance multi-class model fitting} is to determine a set of instances $\{ h_i \}_{i = 1}^{n_i} \subseteq \mathcal{H}_\forall$ and labeling $L \in \Points \rightarrow \mathcal{H}$ assigning each point $p \in \Points$ to an instance from $\{ h_i \}_{i = 1}^{n_i}$ minimizing energy $E$. 

\section{Progressive multi-model fitting}

In this section, a new pipeline is proposed for multi-model fitting.
Before describing each step in depth, a few definitions are discussed. 

\begin{figure}
    \centering
    \includegraphics[width=0.89\columnwidth]{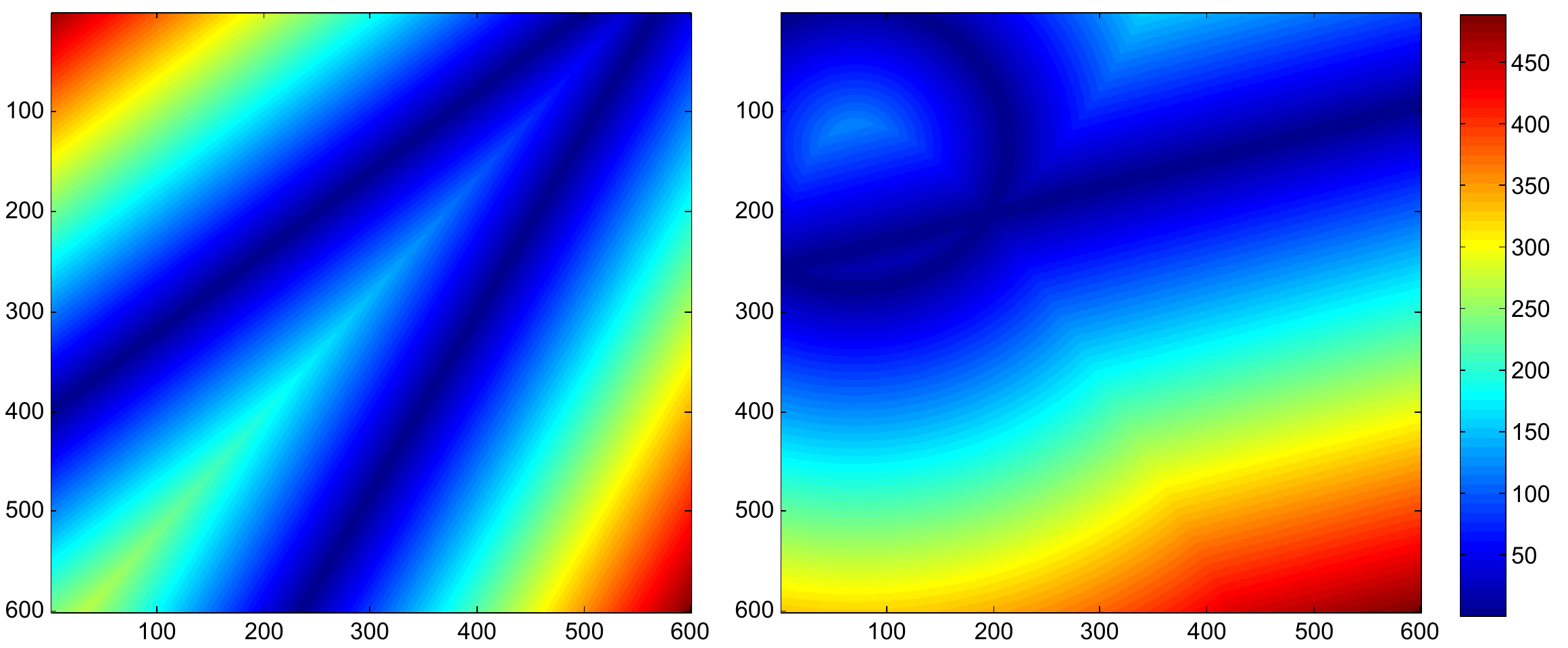}
    \caption{Examples of the compound model, represented by the union of the distance fields, generated by (left) two lines and (right) a line and circle. Best viewed in color. }
    \label{fig:compound_distance_field}
\end{figure}

\begin{defn}%[Putative model instance]
	A {\it putative model instance}  $\Instances_p$ is temporary, generated by the proposal engine, not activated to take part in the optimization procedure.
\end{defn}

\begin{defn}%[Activate model instance]
	An {\it active model instance} $\Instances_a$ is an instance whose parameters and support are  updated in the optimization procedure. 
\end{defn}

\begin{defn}%[Compound model instance]
    {\it Compound model instance}.
	Given a set of activate instances $\{ \Instances_{a,i} \}_{i = 1}^{n_{a}}$, where $n_{a}$ is the cardinality of the set, 
	the compound model instance $\CompoundModel \in \mathcal{H}_\forall$ is defined as the union of the distance fields each generated by an individual active instance $\Instances_{a,i}$ ($i \in [1, n_{a}]$). The distance of a point $\Point$ from $\CompoundModel$ is $\DistanceFunction(\CompoundModel, \Point) = \min_{i = 1}^{n_{a}} \DistanceFunction(\Instances_{a,i}, \Point)$.
\end{defn}  

\noindent
Examples of compound instances are shown in Fig.~\ref{fig:compound_distance_field}. The color codes the distance (blue -- close, red -- far). The left plot shows the union of the distance fields generated by two lines. The right one shows that of a circle and a line. 
\begin{defn}
    \textit{Proposal engine} $\Sigma : \CompoundModel \times \mathcal{P}^* \to \mathcal{H}_\forall^*$ is a function generating a putative instance from the data, using the compound model. Operator $^*$ denotes the power set.  
\end{defn}

\noindent
Therefore, function $\Sigma$ gets the compound model and the set of points and outputs one or multiple proposals.

\begin{defn}
    \textit{Multi-instance optimization} procedure $\Theta : \{ \Instances_{a,i} \}_{i = 1}^{n_{a}} \times \{ \Instances_p \}_{i = 1}^{n_p} \times \mathcal{P}^* \to \{ \widehat{\Instances}_{a,i} \}_{i = 1}^{\widehat{n}_a} \times L$ is a function getting the active, the putative instances and the data as input. It returns a set of active instances and a labeling $L : \mathcal{P}^* \to \Instances_a^*$ which assigns each point to a single active instance. 
\end{defn}
\noindent
Function $\Theta$ gets the active instances $\{ \Instances_{a,i} \}_{i = 1}^{n_{a}}$, set of proposals $\{ \Instances_p \}_{i = 1}^{n_p}$ and input points. It returns the optimized active instances and the labeling. Putative instances can be activated or rejected. 
Also, activate instances may be deactivated and removed here.
The set of active instances always contain an instance of the outlier class. 

\subsection{Proposal engine}
 
The \textit{proposal engine} proposes a yet unseen instance in every iteration. 
For the engine, we choose a recent variant of RANSAC~\cite{fischler1981random}: Graph-Cut RANSAC~\cite{barath2018graph} since it is state-of-the-art and its implementation is publicly available\footnote{\url{https://github.com/danini/graph-cut-ransac}}.
Due to assuming local structures, similarly as in \cite{barath2018multix}, we choose NAPSAC~\cite{nasuto2002napsac} to be the sampler inside GC-RANSAC.
The main objective of the proposal engine is to propose unseen instances, i.e.\ the ones which are possibly not among to active ones in $\{ \Instances_{a,i} \}_{i = 1}^{n_{a}}$. 
A straightforward choice for achieving this goal is to prefer instances having a reasonable number of \textit{points not shared with the compound instance} $\CompoundModel$.  
Therefore, we propose a new quality function $\widehat{Q}: \mathcal{H}_\forall \times \mathcal{H}_\forall^* \times \mathcal{P}^* \times \mathbb{R} \to \mathbb{R}$, for RANSAC, measuring the score of a model instance originating from $\mathcal{H}_\forall$ w.r.t.\ the compound instance (from $\mathcal{H}_\forall^*$), data and a manually set threshold. 
For the sake of easier understanding, we start from one of the simplest quality functions, i.e.\ the inlier counting of RANSAC.
It is as follows:
$
    Q_\text{R}(\Instances, \Points, \epsilon) = \sum_{\Point \in \Points} [\DistanceFunction(\Instances, \Point) < \epsilon],
$
where $[.]$ is the Iverson bracket which is equal to one if the condition inside holds and zero otherwise. Based on $Q_\text{R}$, the modified quality function which takes $\CompoundModel$ into account and, thus, does not count the shared points is the following:
\begin{equation}
    \widehat{Q}_\text{R}(\Instances, \Points, \epsilon) = \sum_{\Point \in \Points} [\DistanceFunction(\Instances, \Point) < \epsilon \wedge \DistanceFunction(\CompoundModel, \Point) \geq \epsilon].
\end{equation}
The condition inside holds if and only if the distance of point $\Point$ from instance $\Instances$ is smaller than $\epsilon$ and, at the same time, its distance from the compound model $\CompoundModel$ is greater than $\epsilon$. 
Therefore, $\widehat{Q}_\text{R}$ counts the points which are not inliers of the compound instance but inliers of the new one.

It nevertheless turned out that, in practice, the truncated quality function of MSAC~\cite{torr2002bayesian} is superior to the inlier counting of RANSAC in terms of accuracy and sensitivity of the user-defined threshold. It is as follows:
\begin{equation}
    Q_\text{M}(\Instances, \Points, \epsilon) = |\Points| - \sum_{\Point \in \Points} \min\left(1, \frac{\DistanceFunction(\Instances, \Point)^2}{\gamma(\epsilon)^2}\right), \; \gamma(\epsilon) = \frac{3}{2} \epsilon.
\end{equation}
Considering the previously described objective, $Q_\text{M}$ is modified as follows to reduce the score of the points which are shared with the compound model: 
\begin{align}
\label{eq:new_quality_function}
    \begin{split}
        \widehat{Q}_\text{M}(\Instances, \Points, \epsilon) = |\Points| - \\
        \sum_{\Point \in \Points} \min\left(1, \max\left( \frac{\DistanceFunction(\Instances, \Point)^2}{\gamma(\epsilon)^2}, 1 - \frac{\DistanceFunction(\CompoundModel, \Point)^2}{\gamma(\epsilon)^2}\right)\right).
    \end{split}
\end{align}
Consequently, for point $\Point$, the implied score is zero (i) if $\Point$ is close to both the proposal and the compound instances, and (ii) if it is far from the proposal.

Summarizing this section, we apply GC-RANSAC as a proposal engine with NAPSAC sampling and quality function $\widehat{Q}_\text{M}$ to propose instances one-by-one.
 
\subsection{Proposal validation}

The \textit{validation} is an intermediate step between the proposal and the optimization to decide if an instance should be involved in the optimization.
To do so, an instance-to-instance distance has to be defined measuring the similarity of the proposed and compound instances. 
If the distance is low, the proposal is likely to be an already seen one and, thus, is unnecessary to optimize. 
In~\cite{barath2018multix}, the instances are represented by problem-specific sequences of points and 
This approach, leads to the question of how to represent models by points. 
In general, the answer is not trivial and the representation affects the outcome significantly.
There, is a straightforward solution for representing instances by point sets.
In~\cite{toldo2008robust}, the models are described by their preference sets, and the similarity of two instances is defined via their Jaccard score. 
The preference set of instance $\Instances$ is $\mathbf{P}_\Instances \in \{0, 1\}^{|\Points|}$, where $\mathbf{P}_{\Instances,j}$ is one if the $j$th point is an inlier of $\Instances$, otherwise zero ($j \in [1, |\Points|]$).
The proposed criterion of accepting a putative instance is
\begin{equation}
    \text{J}(\Instances, \CompoundModel) = \frac{|\mathbf{P}_{\Instances} \cap \mathbf{P}_{\CompoundModel}|}{|\mathbf{P}_{\Instances} \cup \mathbf{P}_{\CompoundModel}|} > \epsilon_{\text{S}},
\end{equation}
where $\text{J}$ holds if the Jaccard similarity is higher than a manually set threshold $\epsilon_{\text{S}} \in [0, 1]$ and $\text{J}$ is false otherwise. 
%If $\text{J}(\Instances, \CompoundModel) = 0$, the distances of the points from the proposal and the compound instances are the same. 
%In case of $\text{T}(\Instances, \CompoundModel) = 1$, they have no points shared.
We choose Jaccard similarity instead of Tanimoto distance~\cite{tanimoto1958elementary,magri2014t} since representing the instances by sets offers a straightforward way of speeding up the procedure.

Computing the Jaccard similarity in case of having thousands of points is a computationally demanding operation. 
Luckily, we are mostly interested in recognizing if the overlap of the instances is significant. 
The min-hash algorithm~\cite{broder1997resemblance} is a straightforward choice for making the processing time of the similarity calculation fast and independent on the number of points.
\textit{Therefore, the validation step runs in constant time.}

\subsection{Multi-instance optimization}

The objective of this step is to optimize the set of active model instances whenever a new putative one comes and to decide if this new instance shall be activated or should be rejected.
% This step of the pipeline is where state-of-the-art multi-model fitting algorithms are involved as they focus mostly on solving this optimization problem.
%
Due to aiming at the most general case, i.e.\ having multiple classes, there are just a few algorithms~\cite{isack2012energy,delong2012minimizing,pham2014interacting,barath2018multix} which can approach the problem without requiring non-trivial modifications.
These algorithms are based on labeling energy-minimization.
In general, the major issue of labeling algorithms is their computational complexity, especially, in the case of large label space. 
In our case, due to proposing instances one-by-one, the label space is always kept small and, therefore, the time spent on the labeling is not significant. 

Multi-X~\cite{barath2018multix} could be a justifiable choice. 
However, in our case, it is simplified to the PEARL algorithm~\cite{isack2012energy,delong2012fast} since its major contribution is a move in the label space replacing a label set with the corresponding density mode. 
When having just a few labels, this move is not needed. 
Thus, we choose PEARL as the optimization procedure.\footnote{\url{https://github.com/nsubtil/gco-v3.0}}  
%Note that, in case of having geometric priors, e.g.\ the objective is to find orthogonal planes, MFIGP is also a suitable choice~\cite{pham2014interacting}.

\subsection{Termination criterion}

In this section, we propose a criterion to determine when to stop with proposing new instances. The adaptive termination criterion of RANSAC is based on 
\begin{equation}
    1 - \mu \leq \left(1 - \left(|\mathcal{I}| / |\Points| \right)^m \right)^k,
    \label{eq:orig_termination}
\end{equation}
where $\mu$ is the required confidence in the results typically set to $0.95$ or $0.99$; $k$ is the number of iterations; $m$ is the size of the minimal sample; $|\mathcal{I}|$ and $|\Points|$ are the number of inliers and points, respectively. 
In RANSAC, Eq.~\ref{eq:orig_termination} is formulated to determine $k$, i.e.\ the number of iterations required, using the current inlier ratio. We, instead, formulate Eq.~\ref{eq:orig_termination} to have an estimate of the maximum number of inliers independent on the compound model as follows:
$
    (|\Points| - |\CompoundModel|) \sqrt[m]{1 - \sqrt[k]{1 - \mu}} \geq |\mathcal{I}|
$
where $|\CompoundModel|$ is the inlier number of the compound model.
Therefore, $\overline{I}(\Points, \CompoundModel, m, k, \mu) = (|\Points| - |\CompoundModel|) \sqrt[m]{1 - \sqrt[k]{1 - \mu}}$ is an upper limit for the inlier number of a not yet found instance with confidence $\mu$ in the $k$th iteration. 

It can be easily seen that to distinguish an instance, at least $m + 1$ points have to support it, where $m$ is the size of a minimal sample. 
Therefore, the algorithm updates $\overline{I}$ in every iteration and terminates if $\overline{I} < m + 1$.
\textit{This constraint guaranties} that when the algorithm terminates, the probability of having an unseen model with at least $m + 1$ inliers is smaller than $1 - \mu$.

Note that in practice, it is more convenient to set this limit on the basis of the optimization. For instance, if the optimization does not accept instances having fewer than $20$ inliers, it does not make sense to propose ones with fewer. 

\section{Experimental results}

In this section, we evaluate the proposed Prog-X method on various computer vision problems that are: 2D line, homography, two-view motion, motion estimation, and simultaneous plane and cylinder fitting. 
% The parameters of the evaluated methods are tuned for obtaining the best average accuracy over five runs on each dataset. 
The reported results are the averages over five runs and obtained by using fixed parameters. 
See Table~\ref{tab:parameter_table} for the parameters of Prog-X.

\subsection{Synthesized tests}

\begin{figure*}
    \centering
    \begin{subfigure}[t]{0.9\textwidth}
   	 	\centering
        \includegraphics[width=0.14\columnwidth]{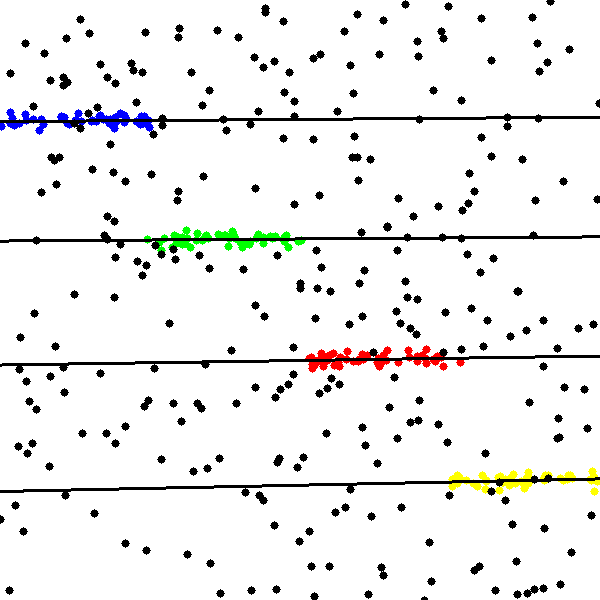}\phantom{xx}
        \includegraphics[width=0.14\columnwidth]{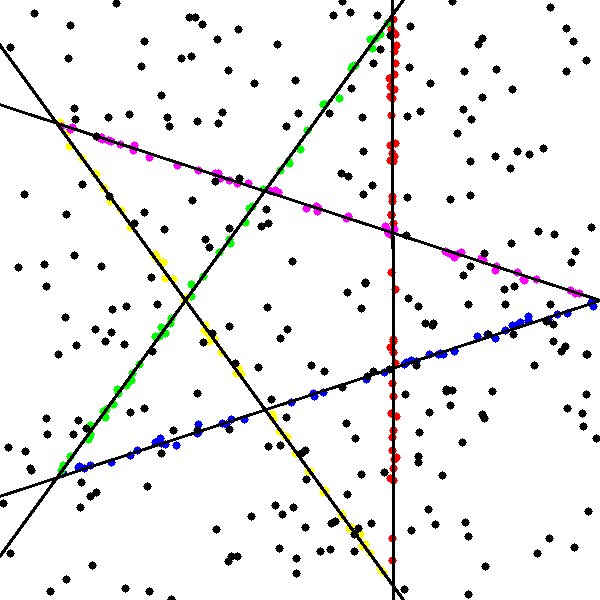}\phantom{xx}
        \includegraphics[width=0.14\columnwidth]{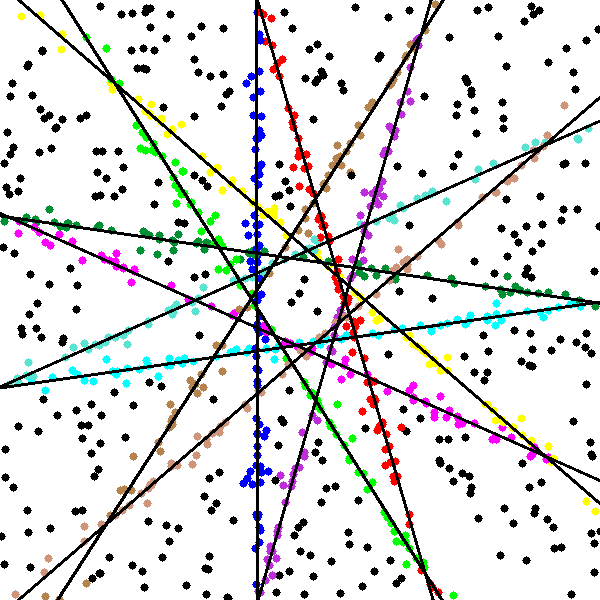}
        \caption{Ground truth instances and clustering coded by color}
        \label{fig:gt_star_dataset}
    \end{subfigure}\\
    \begin{subfigure}[t]{0.02\textwidth}
   	 	\centering \vspace{-1.9cm}
   	 	\rot{\fontfamily{cmtt}\selectfont{stair4}}\vspace{1.75cm}
   	 	\rot{\fontfamily{cmtt}\selectfont{star5}}\vspace{1.8cm}
   	 	\rot{\fontfamily{cmtt}\selectfont{star11}}
    \end{subfigure}\hfill
    \begin{subfigure}[t]{0.15\textwidth}
   	 	\centering
        \includegraphics[width=1.0\columnwidth]{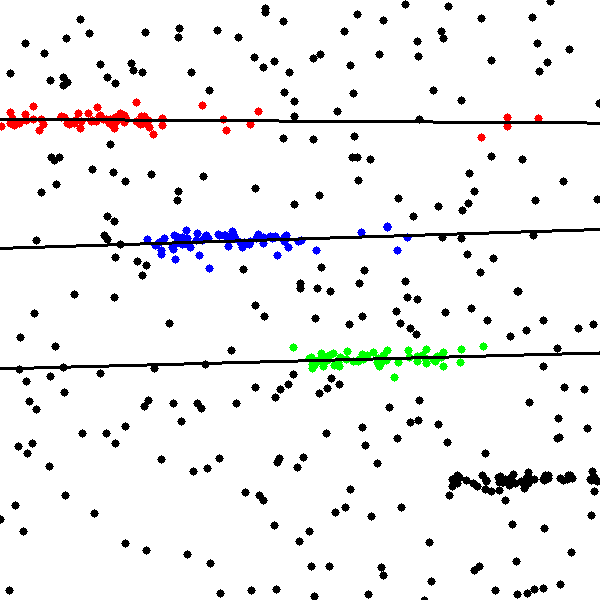}
        \includegraphics[width=1.0\columnwidth]{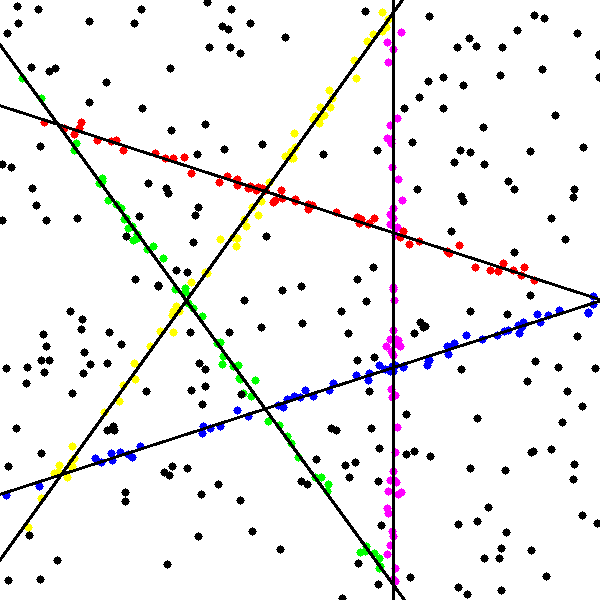}
        \includegraphics[width=1.0\columnwidth]{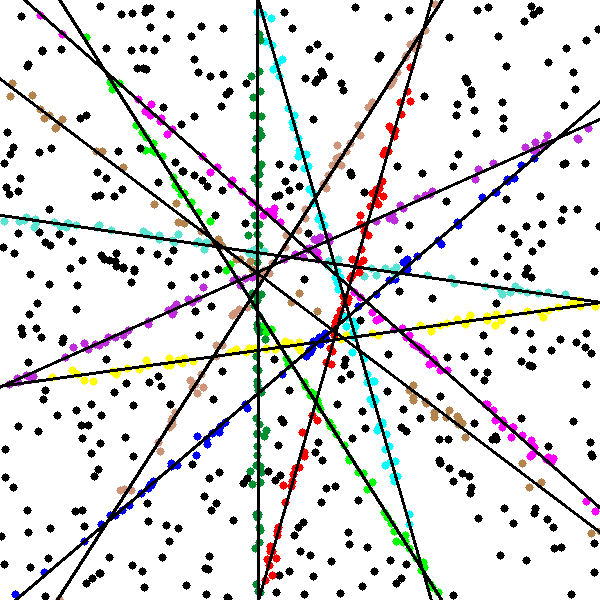}
        \caption{Prog-X}
    \end{subfigure}\hfill
    \begin{subfigure}[t]{0.15\textwidth}
   	 	\centering
        \includegraphics[width=1.0\columnwidth]{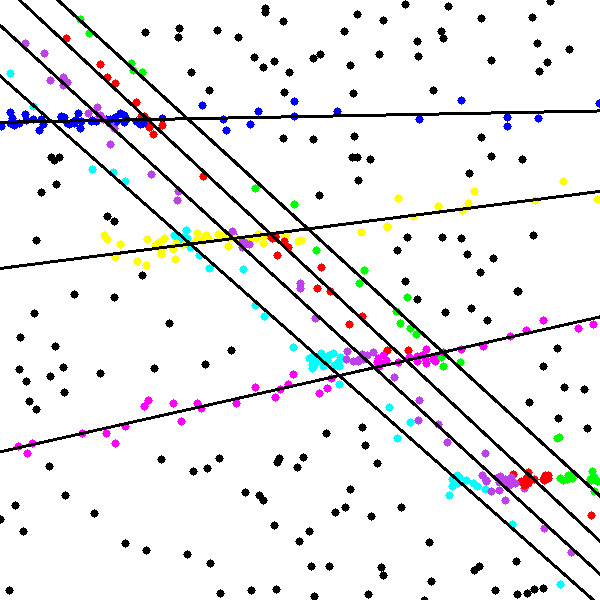}
        \includegraphics[width=1.0\columnwidth]{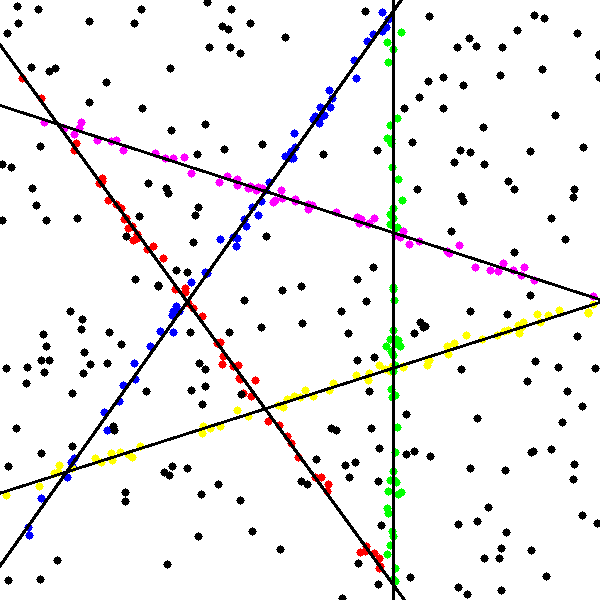}
        \includegraphics[width=1.0\columnwidth]{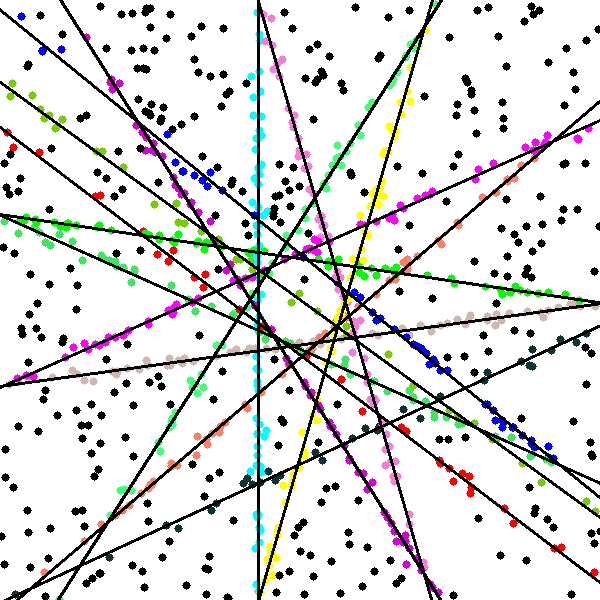}
        \caption{Multi-X}
    \end{subfigure}\hfill
    \begin{subfigure}[t]{0.15\textwidth}
   	 	\centering
        \includegraphics[width=1.0\columnwidth]{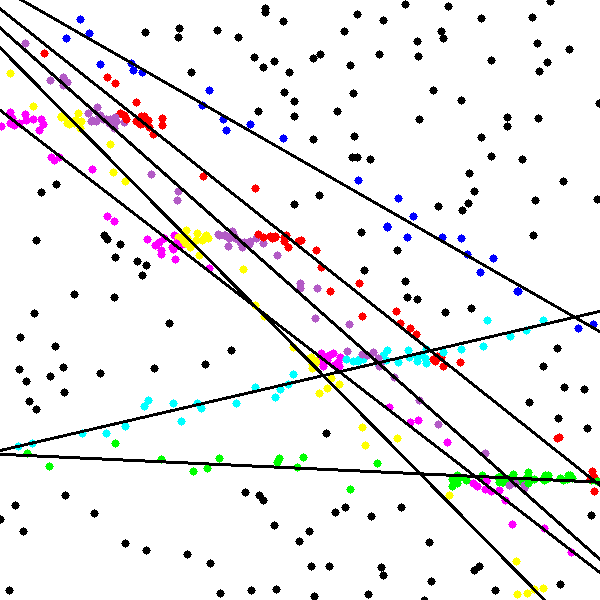}
        \includegraphics[width=1.0\columnwidth]{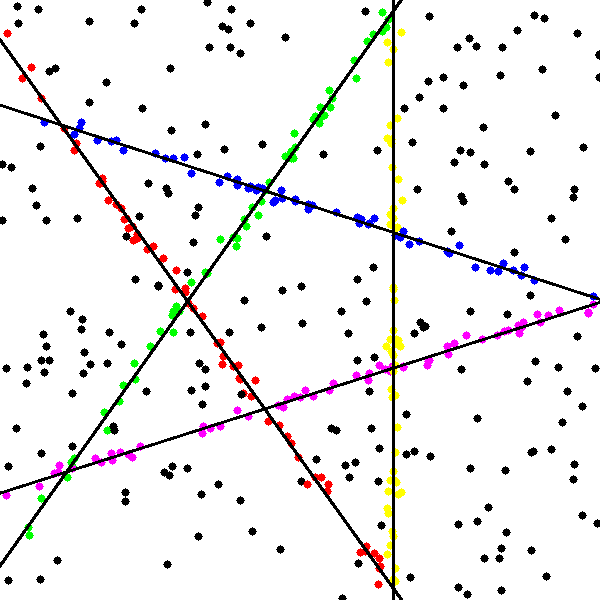}
        \includegraphics[width=1.0\columnwidth]{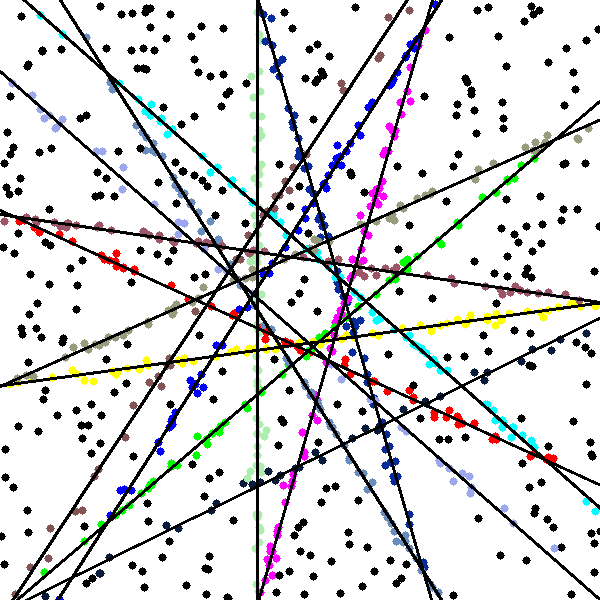}
        \caption{PEARL}
    \end{subfigure}\hfill
    \begin{subfigure}[t]{0.15\textwidth}
   	 	\centering
        \includegraphics[width=1.0\columnwidth]{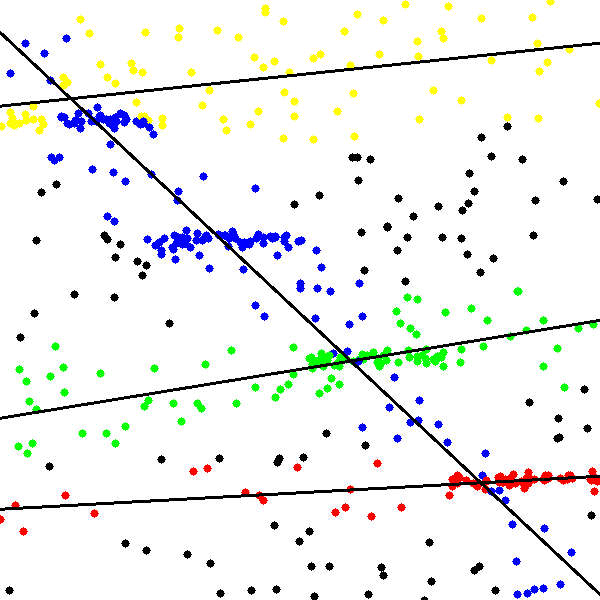}
        \includegraphics[width=1.0\columnwidth]{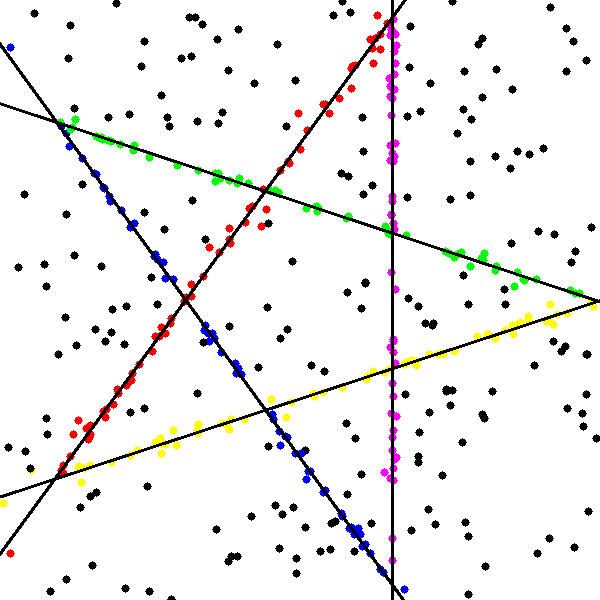}
        \includegraphics[width=1.0\columnwidth]{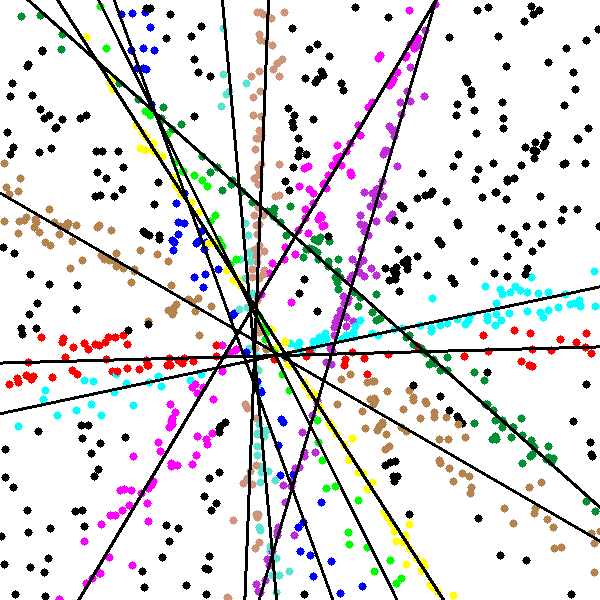}
        \caption{RPA}
    \end{subfigure}\hfill
    \begin{subfigure}[t]{0.15\textwidth}
   	 	\centering
        \includegraphics[width=1.0\columnwidth]{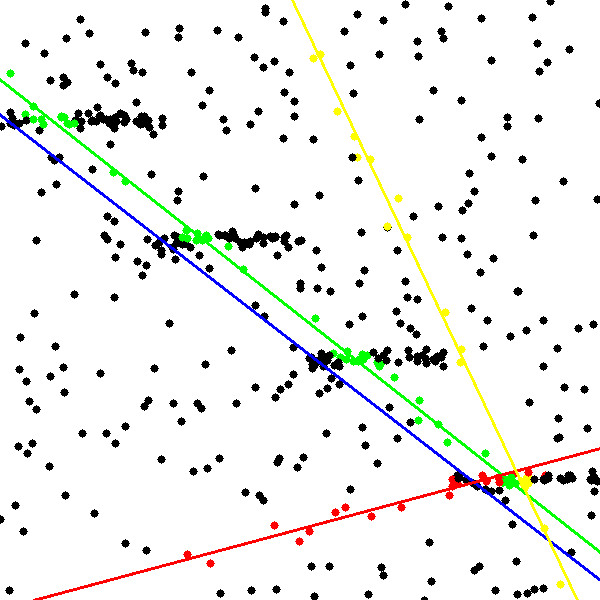}
        \includegraphics[width=1.0\columnwidth]{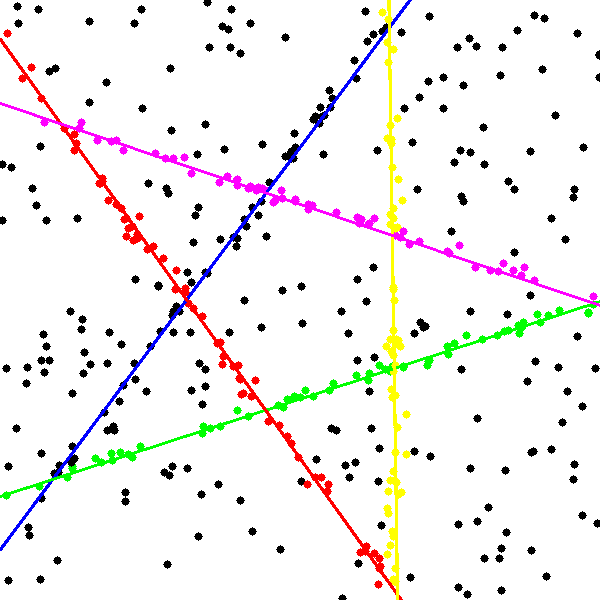}
        \includegraphics[width=1.0\columnwidth]{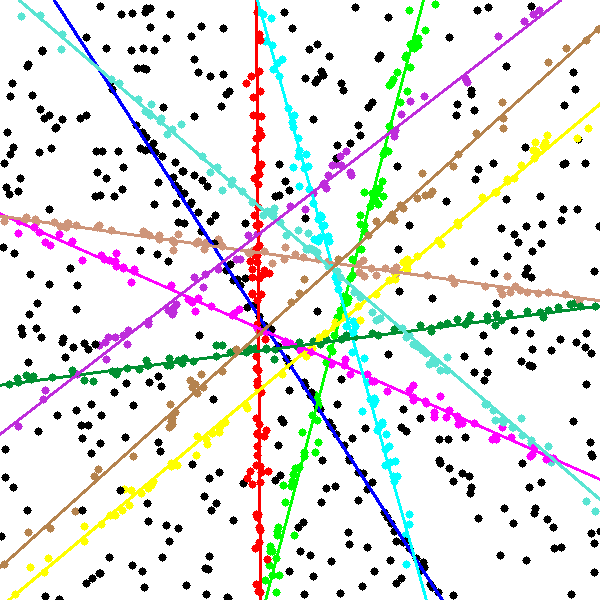}
        \caption{RansaCov}
    \end{subfigure}\hfill
    \begin{subfigure}[t]{0.15\textwidth}
   	 	\centering
        \includegraphics[width=1.0\columnwidth]{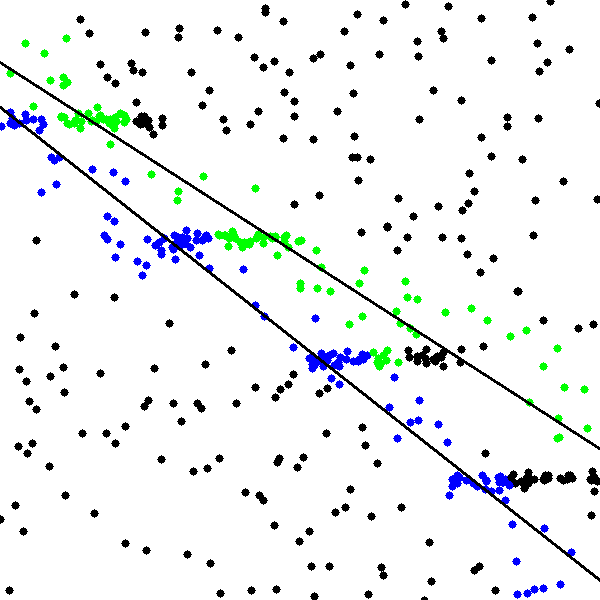}
        \includegraphics[width=1.0\columnwidth]{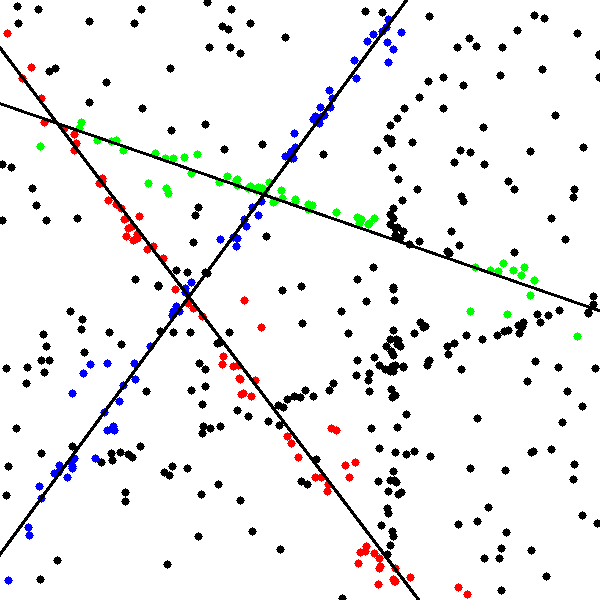}
        \includegraphics[width=1.0\columnwidth]{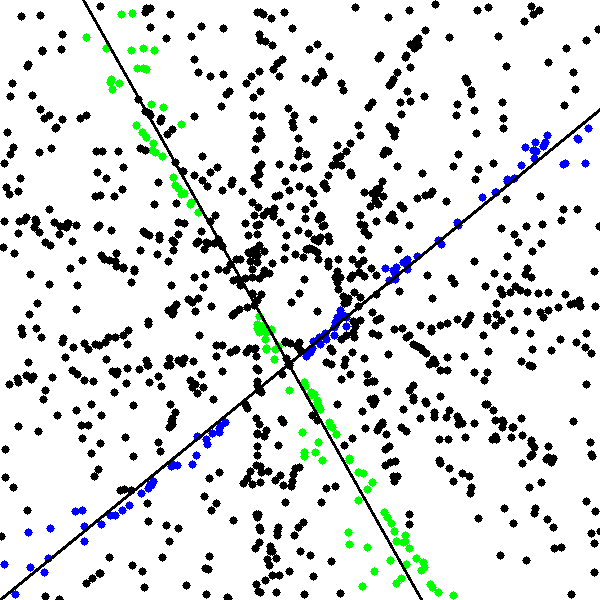}
        \caption{T-Linkage}
    \end{subfigure}
    \caption{ 
    Fitting of multiple 2D lines to synthetic data.
    The \textit{worst results} over five runs on the dataset of~\cite{toldo2008robust} are shown. The points are assigned to lines by color.  
    In (a), the ground truth clusterings are shown. From (b) to (g), the results of the evaluated algorithms are put. 
    Fixed parameters were used for all methods. See Table~\ref{tab:linefitting_table} for the details of the results.  }
    \label{fig:linefitting_visual}
\end{figure*}

In this section Prog-X is tested in a synthetic environment. 
We chose line fitting to 2D points and downloaded the dataset used in~\cite{toldo2008robust}. It consists of three scenes: {\fontfamily{cmtt}\selectfont{stair4}}, {\fontfamily{cmtt}\selectfont{star5}} and {\fontfamily{cmtt}\selectfont{star11}} (see Fig.~\ref{fig:gt_star_dataset}). 
% In {\fontfamily{cmtt}\selectfont{stair4}}, the line segments are fairly short and almost parallel to each other. 
% The difficulty of this scene is that drawing a single line which crosses all the four line segments leads to lower energy if the spatial relationships of the points are not considered.
% In {\fontfamily{cmtt}\selectfont{star5}}, there are five lines forming a star, and in {\fontfamily{cmtt}\selectfont{star11}} there are eleven ones. 
The outlier ratio in all test cases is $0.5$ thus having equal number of inliers and outliers.

\begin{table*}
	\center
	\begin{tabular}{  r | c c | c c | c c | c c | c c | c c  }
    \hline
         & \multicolumn{2}{c|}{\cellcolor{black!5} \textbf{Progressive-X}}  & \multicolumn{2}{c|}{Multi-X}  & \multicolumn{2}{c|}{PEARL}  & \multicolumn{2}{c|}{RPA}  & \multicolumn{2}{c|}{RansaCov}  & \multicolumn{2}{c}{T-Linkage} \\
    \hline
         & \cellcolor{black!5}\# fn & \cellcolor{black!5}\# fp & \# fn & \# fp & \# fn & \# fp & \# fn & \# fp & \# fn & \# fp & \# fn & \# fp \\
    \hline
        {\fontfamily{cmtt}\selectfont{stair4}} & \cellcolor{black!5}\textbf{1} & \cellcolor{black!5}\textbf{0} & \textbf{1} & 4 & 2 & 5 & 2 & 2 & 4 & 4 & 4 & 2 \\
        {\fontfamily{cmtt}\selectfont{star5}} & \cellcolor{black!5}\textbf{0} & \cellcolor{black!5}\textbf{0} & \textbf{0} & \textbf{0} & \textbf{0} & \textbf{0} & \textbf{0} & \textbf{0} & \textbf{0} & \textbf{0} & 2 & \textbf{0} \\
        {\fontfamily{cmtt}\selectfont{star11}} & \cellcolor{black!5}\textbf{0} & \cellcolor{black!5}\textbf{0} & \textbf{0} & 2 & \textbf{0} & 3 & 5 & 5 & 1 & 1 & 9 & \textbf{0} \\
    \hline
        {\fontfamily{cmtt}\selectfont{all}} & \cellcolor{black!5}\textbf{1} & \cellcolor{black!5}\textbf{0} & \textbf{1} & 6 & 2 & 8 & 7 & 7 & 5 & 5 & 15 & 2 \\
    \hline
    \end{tabular}
	\caption{Fitting of multiple 2D lines to synthetic data.
	The \textit{worst results} over five runs on the dataset of~\cite{toldo2008robust} are shown.
	The reported properties are the number of false negative (\# fn) and false positive instances (\# fp). 
	Fixed parameters were used for all methods. 
	The results are visualized in Fig.~\ref{fig:linefitting_visual}. }
    \label{tab:linefitting_table}
\end{table*}

\noindent
\textbf{Comparison of multi-model fitting algorithms.}
We compare Prog-X, Multi-X\footnote{\url{https://github.com/danini/multi-x}},~\cite{barath2018multix} RansaCov~\cite{magri2016multiple}, RPA~\cite{magri2015robust}, T-Linkage\footnote{\url{http://www.diegm.uniud.it/fusiello/demo/}}~\cite{magri2014t}, and PEARL~\cite{boykov2001fast} on line fitting in this sections. 
All methods were applied five times to all scenes with fixed parameters. 
For T-Linkage, RPA and RansaCov, we used the parameters which the authors proposed.
We tuned Prog-X, Multi-X, and PEARL to achieve the most accurate average results. 

The \textit{worst results} in five runs are visualized in Fig.~\ref{fig:linefitting_visual}. 
Plot (a) shows the ground truth lines and point-to-line assignments. 
The points of each cluster are drawn by color. 
The number of false negative, i.e.\ a line which is not found, and false positive, i.e.\ a found line which is not in the ground truth set, instances are reported in Table~\ref{tab:linefitting_table}.
It can be seen, that the proposed method leads to the most accurate results.
It finds all but one lines and does not return false positives. 

\noindent
\textbf{Any-time property.} To demonstrate the any-time property of Prog-X, we applied the methods which minimize an energy function iteration-by-iteration, i.e.\ Prog-X, PEARL~\cite{boykov2001fast} and Multi-X~\cite{barath2018multix}, to the {\fontfamily{cmtt}\selectfont{star11}} scene and reported their states in each iteration. 
All methods were applied five times and, for each, the run with the worst outcome was selected. 

\begin{figure}
    \centering
    \begin{subfigure}[t]{0.45\columnwidth}
        \includegraphics[width=0.99\columnwidth]{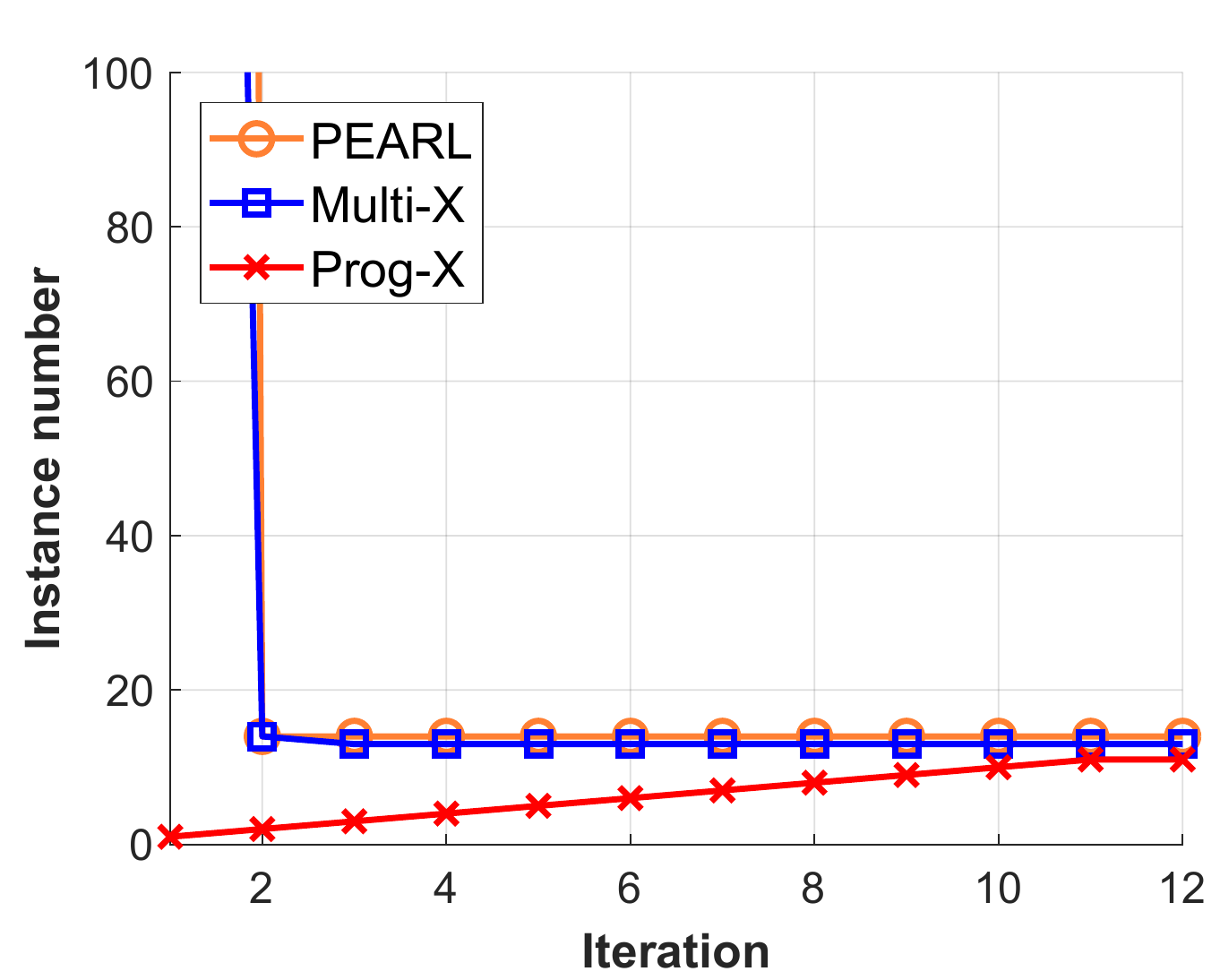}
        \caption{}
        \label{fig:anytime_plots_a}
    \end{subfigure}
    \begin{subfigure}[t]{0.45\columnwidth}
        \includegraphics[width=0.99\columnwidth]{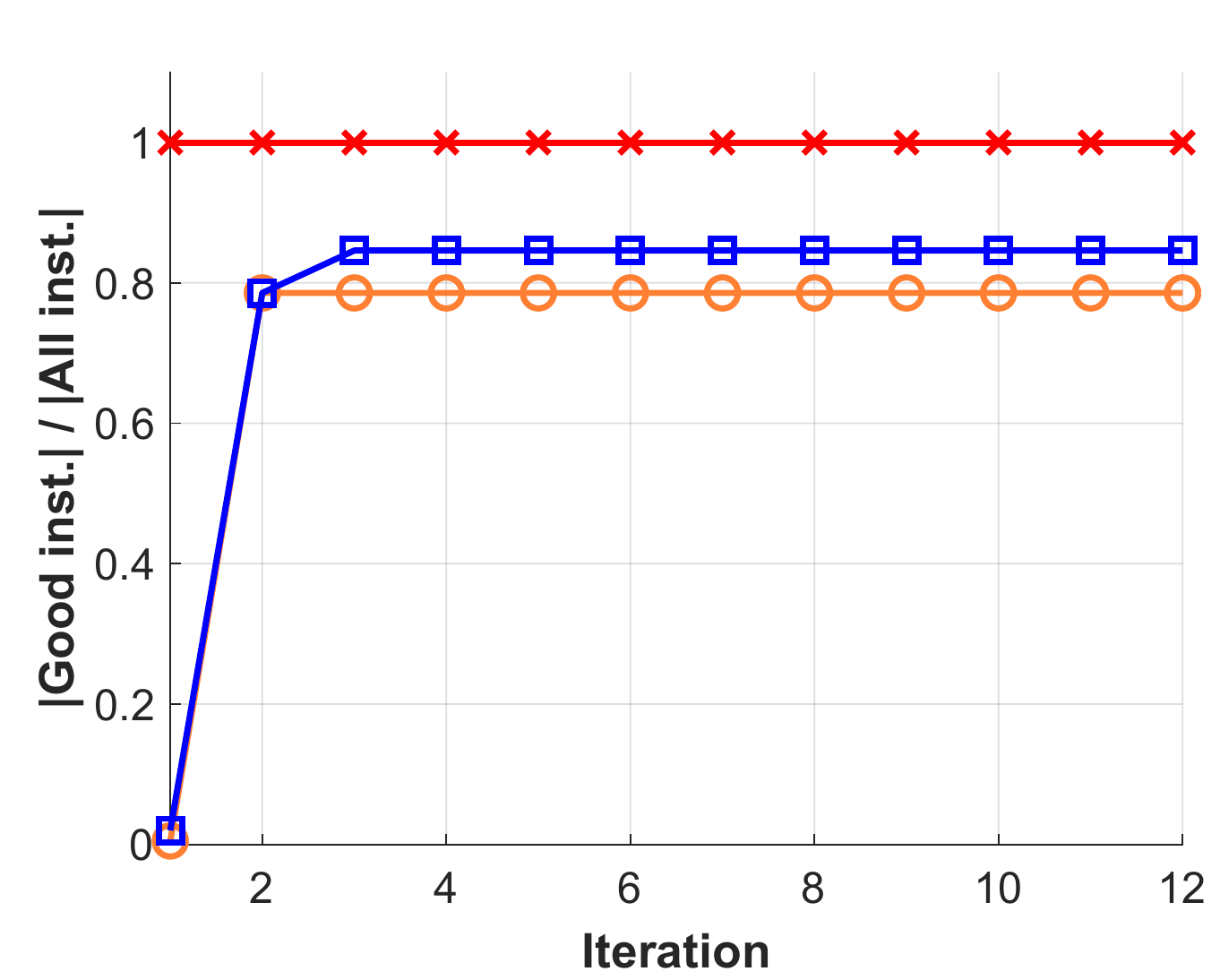}
        \caption{}
    \end{subfigure}
    \begin{subfigure}[t]{0.45\columnwidth}
    \includegraphics[width=0.99\columnwidth]{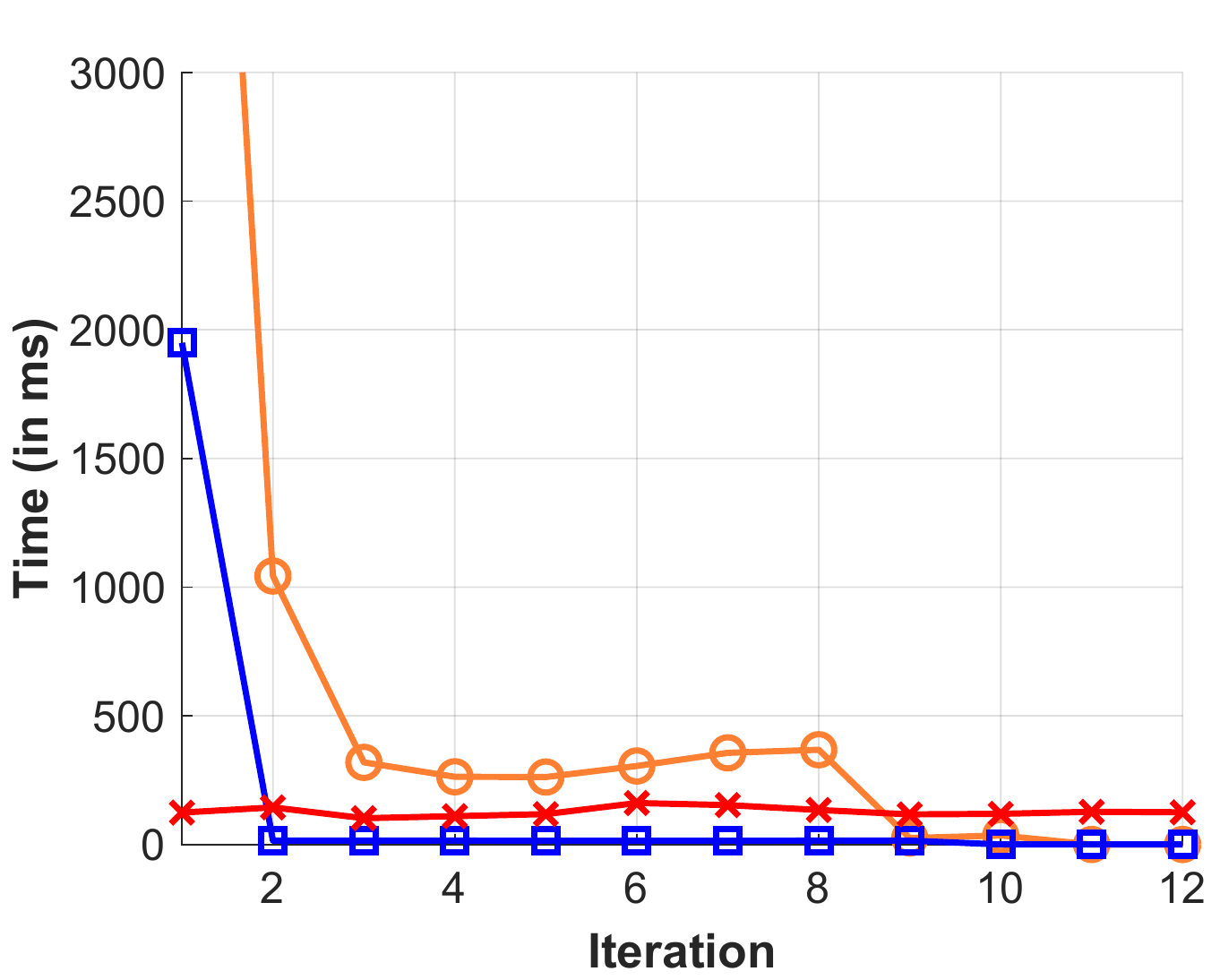}
        \caption{}
    \end{subfigure}
    \begin{subfigure}[t]{0.45\columnwidth}
    \includegraphics[width=0.99\columnwidth]{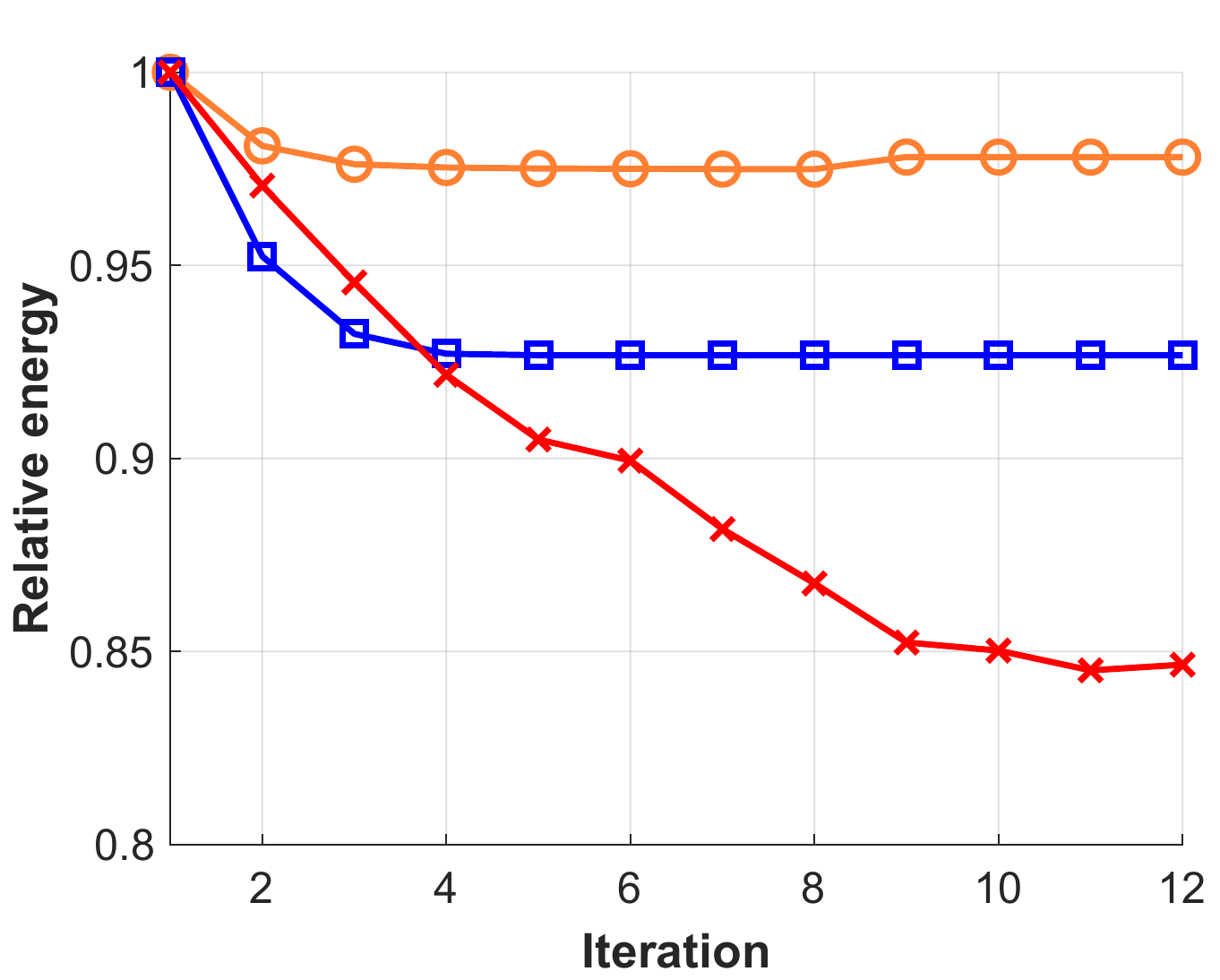}
        \caption{}
    \end{subfigure}
    \caption{\textit{The results iteration-by-iteration.} The compared methods minimizing an energy function iteratively: the proposed Prog-X (red), Multi-X~\cite{barath2018multix} (blue) and PEARL~\cite{boykov2001fast} (orange). 
    \textbf{(a)} The number of stored instances are shown (vertical axis) as a function of the iteration number (horizontal). 
    The values in the $1$st iteration are $2\;200$ for PEARL and $550$ for Multi-X. 
    \textbf{(b)} The ratio (vertical) of the number of ground truth models covered by an instance and the number of all instances stored are reported. 
    \textbf{(c)} The processing times (vertical; in ms) are shown for each iteration (horizontal). 
    The times in the $1$st iteration are $6\;904$ ms for PEARL and $1\;949$ ms for Multi-X. 
    In total, PEARL required $9\;880$ ms, Multi-X $2\;065$ ms and Prog-X $1\;534$ ms. 
    \textbf{(d)} The energy divided by the energy in the $1$st iteration (horizontal) is is shown. 
  }
    \label{fig:anytime_plots}
\end{figure}

Fig.~\ref{fig:anytime_plots} plots the evolution of the properties iteration-by-iteration.
In (\textbf{a}), the numbers of stored instances (vertical axis) are plotted as the function of the iteration number (horizontal). 
It can be seen that PEARL and Multi-X have an unnecessarily high number of instances stored in the first iteration. 
Then, from the second one, it drops significantly and remains almost constant. 
For Prog-X, the number of instances increases by one in each iteration. 
The ground truth number was $11$ in this scene. 
In (\textbf{b}), it is simulated what would happen if a method is stopped before it terminates.
In this case, all methods have a set of instances stored. 
In the plot, the ratio (vertical axis) of the number of desired instances kept and the number of all stored instances is shown as the function of the iteration number (horizontal).
A desired instance is one which overlaps with a ground truth one. 
A ground truth instance can be covered only by one instance. 
The number of desired instances is, thus, at most the number of ground truth models in the data. 
It can be seen that PEARL and Multi-X are significantly worse than Prog-X in the first iteration since the number of stored instances is far bigger than the number of desired ones. 
Even if all the ground truth ones are covered, there are many false positives and, thus, the instances are not usable without further optimization.
This ratio for Prog-X is one in all iterations and, as a consequence, \textit{it can be stopped any time and still returns solely desired instances}.
In (\textbf{c}), the processing times in milliseconds (vertical axis) are reported for each iteration (horizontal). PEARL and Multi-X spend seconds in the first iteration and, then, their processing times drop significantly. This is the expected behavior knowing that their most time-consuming operation is the $\alpha$-expansion algorithm which has to manage a fairly large label space in the first iterations. 
For Prog-X, the processing time is almost constant since it alternates between the proposal and optimization step with slightly increasing label space. 
\textbf{Summarizing} (a--c), PEARL and Multi-X are not any-time algorithms. 
To (a) and (b), they can be stopped after the first iteration without significant deterioration in the accuracy. However, to (c), the first iteration requires more processing time than the rest in total.  
Prog-X can be interrupted at any time; all of the stored instances cover ground truth ones.

Plot (d) shows the change of the energy (vertical axis) iteration-by-iteration (horizontal). 
%Even though all methods use the same energy function, their parameters are different. Thus, for every method, 
We divided the energy in each iteration by the energy in the first one. 
It can be seen that Prog-X leads to a significantly higher reduction in the energy than the competitor algorithms. 

\begin{figure}
    \centering
    \begin{subfigure}[t]{0.45\columnwidth}
        \includegraphics[width=0.99\columnwidth]{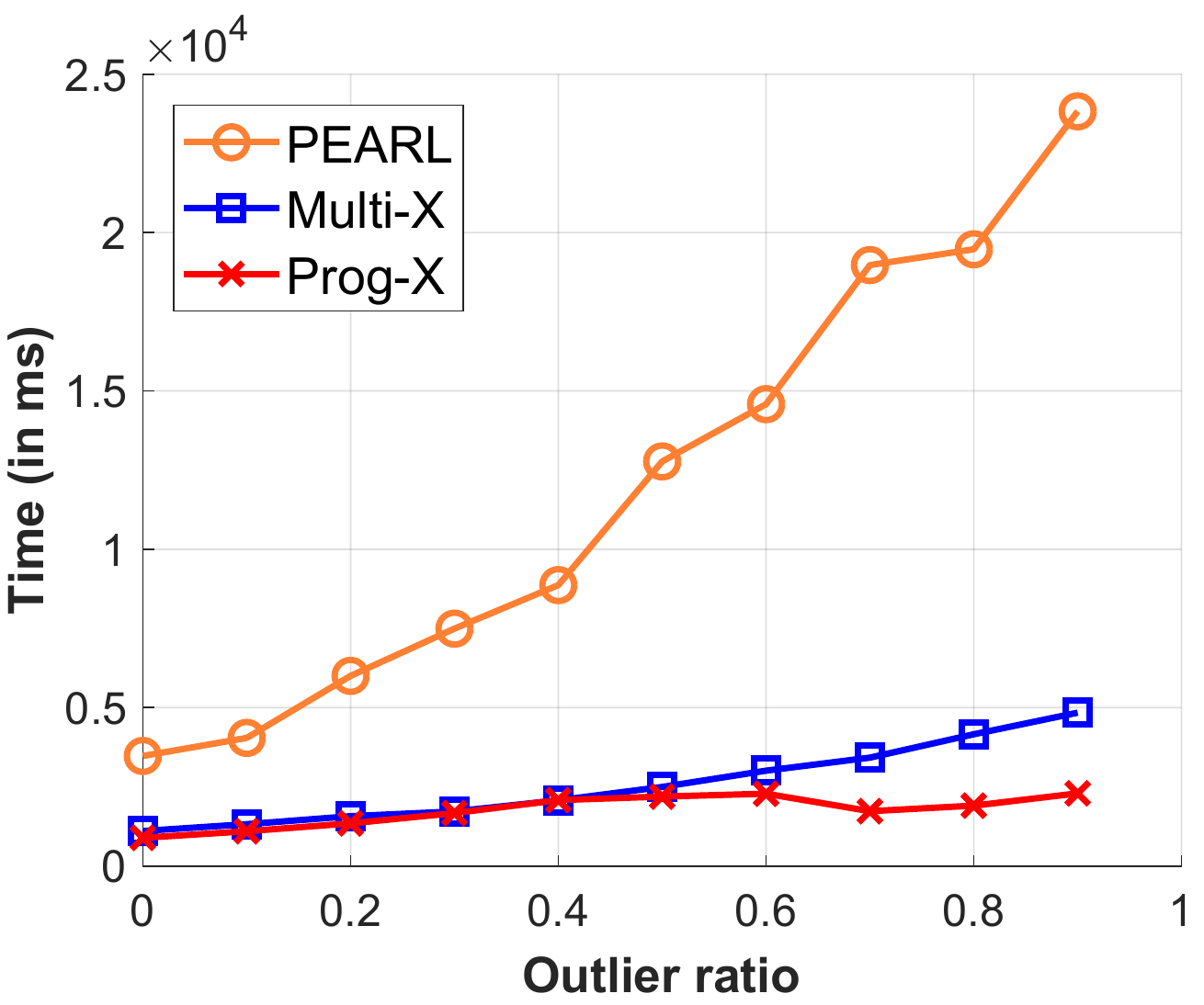}
        \caption{}
    \end{subfigure}
    \begin{subfigure}[t]{0.45\columnwidth}
        \includegraphics[width=0.99\columnwidth]{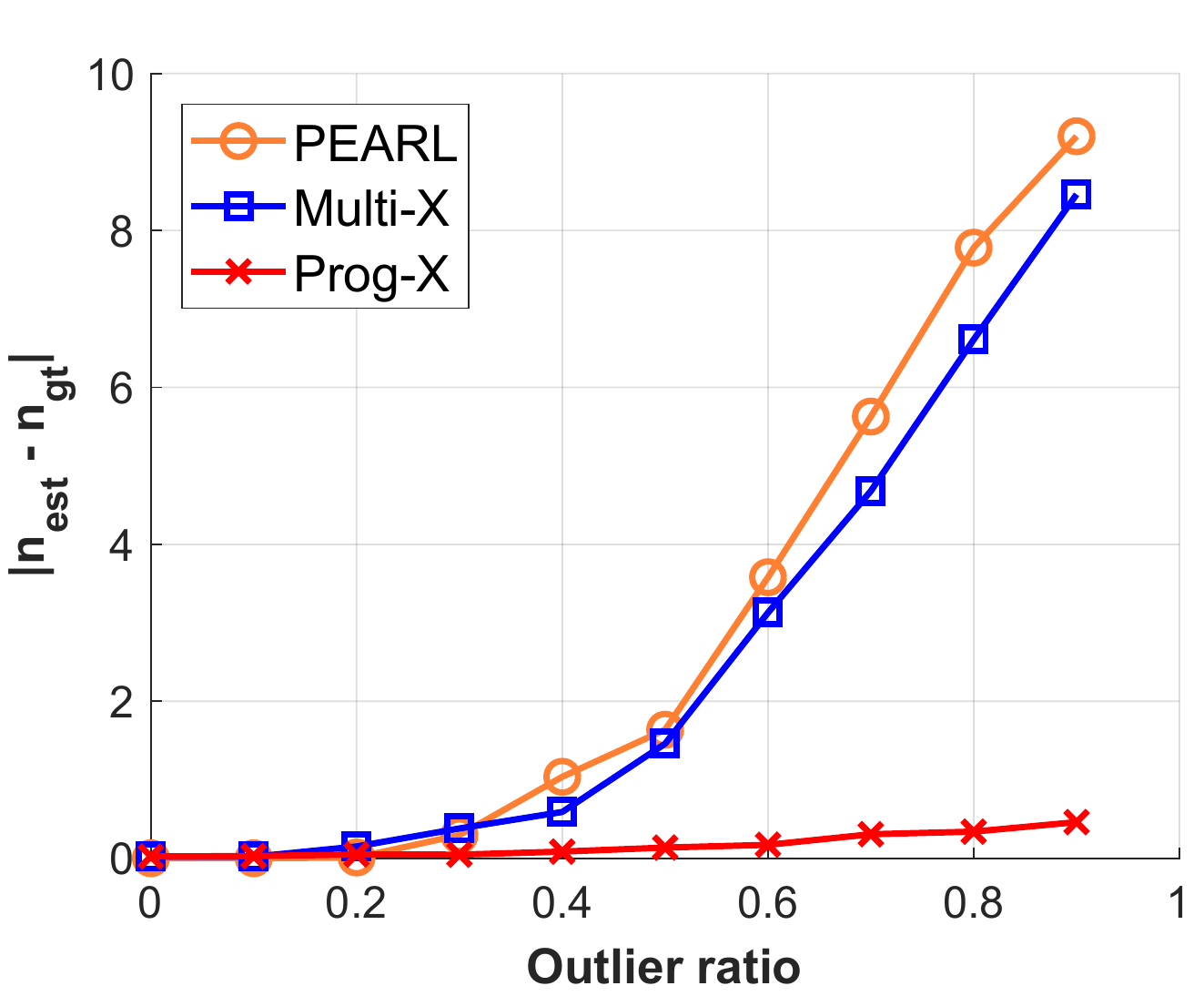}
        \caption{}
    \end{subfigure}
    \caption{(\textbf{a}) The average (of $1\;000$ runs) processing time (in msecs; vertical axis) and (\textbf{b}) the difference of ground truth and returned instance numbers (vertical) plotted as the function of the outlier ratio (horizontal) on {\fontfamily{cmtt}\selectfont{star11}}. In each run, random, uniformly distributed, outliers are generated and the coordinates of the original noisy inliers are perturbed by zero-mean Gaussian-noise with $\sigma = 1.0$ pixels.  
    }
    \label{fig:increasing_outlier_ratio}
\end{figure}

\noindent
\textbf{Comparison with increasing outlier ratio.}
To evaluate, how the outlier ratio affects the outcome of multi-model fitting algorithms, we first kept solely the noisy inliers from the {\fontfamily{cmtt}\selectfont{star11}} scene. Then outliers, i.e.\ points uniformly distributed in the scene, were added to achieve a given outlier ratio $\nu$. 
Fig.~\ref{fig:increasing_outlier_ratio} reports (a) the processing time (in milliseconds; vertical axis) and (b) the difference between the ground truth and the obtained instance number (vertical) as the function of the outlier ratio (horizontal).
It can be seen that the proposed Prog-X leads to similar processing time as Multi-X but the returned number of instances is significantly closer to the desired one. 
Consequently, it is the least sensitive to the outlier ratio. 

\begin{table}
	\center
	\resizebox{0.999\columnwidth}{!}{
	\begin{tabular}{  r | c | c | c | c | c  }
    \hline
       \cellcolor{black!10} & \cellcolor{black!10}$\mu$ & \cellcolor{black!10}$\epsilon_\text{S}$ & \cellcolor{black!10}$w_s$ & \cellcolor{black!10}$\epsilon$ & \cellcolor{black!10}$w_l$ \\
    \hline
        Two-view motions & \multirow{4}{*}{0.95} & \multirow{4}{*}{0.1} & \multirow{4}{*}{0.15} & 1.6 & 43.0 \\
        Homographies & & & & 2.9 & 112.5 \\
        Motions & & & & $5 * 10^{-4}$ & $2.25 * 10^{-6}$  \\
        Planes, cylinders & & & & $0.5$ & $0.035$  \\
    \hline
    \end{tabular}}
	\caption{ Parameters of Prog-X used for each problem (rows). The RANSAC confidence ($\mu$); minimal Jaccard-distance for passing the instance validation ($\epsilon_\text{T}$); inlier-outlier threshold ($\epsilon$); label cost ($w_l$); and the weight of the spatial coherence term ($w_s$) are reported.}
    \label{tab:parameter_table}
\end{table}

\begin{figure*}
    \centering
    \includegraphics[width=1.00\columnwidth]{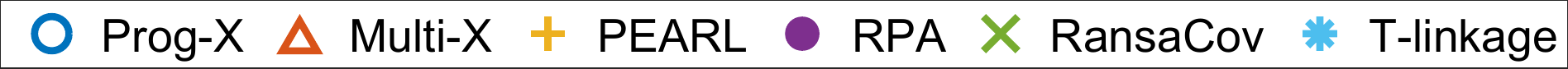}\\[0.2cm]
    \includegraphics[width=1.90\columnwidth]{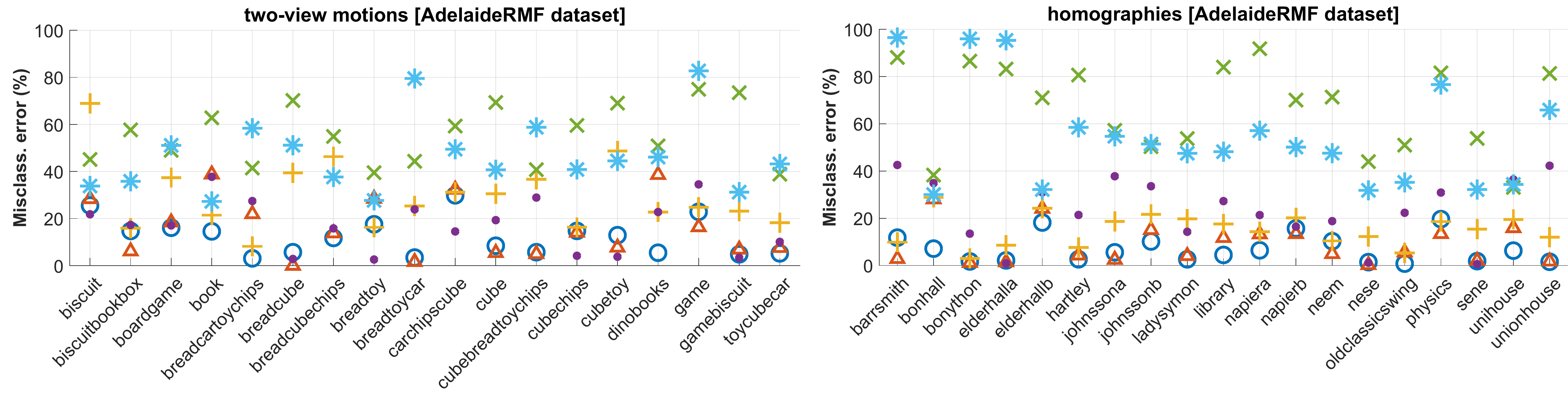}
    \includegraphics[width=1.51\columnwidth]{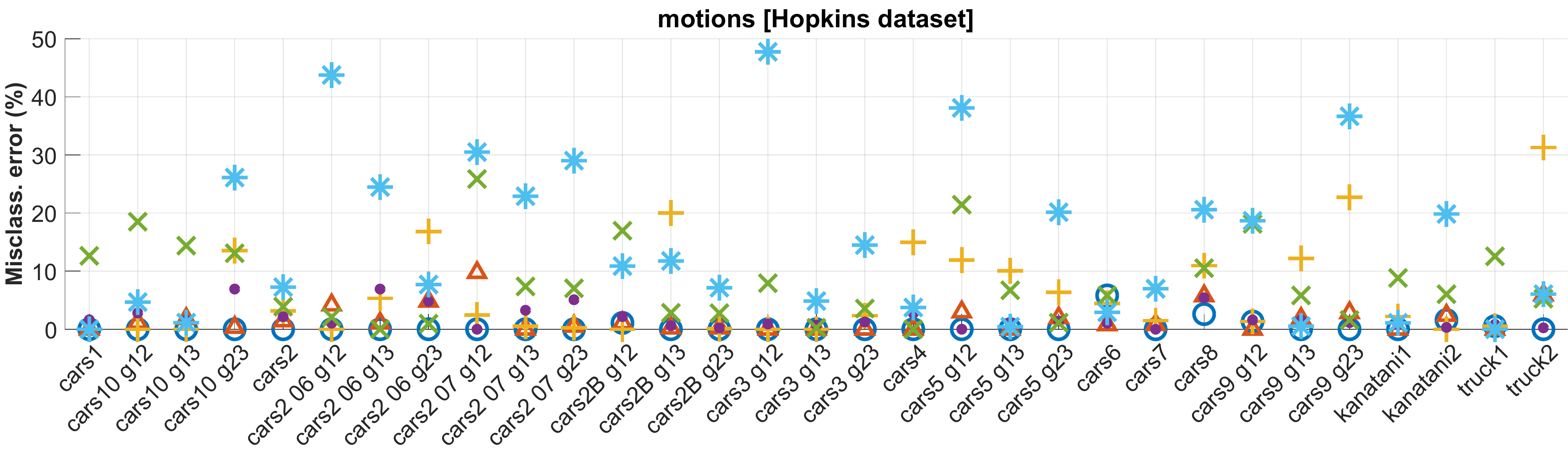}
    \caption{The average ME (vertical axis; in percentage; avg.\ of $5$ runs) of multi-model estimators applied to problems: two-view motion (AdelaideRMF \textbf{F} dataset; top-left), multi-homography (AdelaideRMF \textbf{H} dataset; top-right), multi-motion (Hopkins dataset, {\fontfamily{cmtt}\selectfont{traffic}} sequences; bottom) fitting. All methods used fixed parameters. Summarized results are in Table~\ref{tab:summary_table}.} 
    \label{fig:detailed_results}
\end{figure*}

\begin{table*}
	\center
	\resizebox{0.99\textwidth}{!}{
	\begin{tabular}{ r | c c c | c c c | c c c | c c c  }
    \hline
   		\cellcolor{black!10} & \multicolumn{3}{ c | }{ \cellcolor{black!10}Two-view motions} & \multicolumn{3}{ c | }{ \cellcolor{black!10}Homographies} & \multicolumn{3}{ c | }{ \cellcolor{black!10}Motions } & \multicolumn{3}{ c }{ \cellcolor{black!10}Planes and cylinders } \\
    \hline
   		 & \multicolumn{3}{ c | }{ 18 scenes } & \multicolumn{3}{ c | }{ 19 scenes } & \multicolumn{3}{ c  }{ 155 scenes }  & \multicolumn{3}{ c  }{ 8 scenes } \\
    \hline
   		 & avg. & std. & time & avg. & std. & time & avg. & std. & time & avg. & std. & time \\
    \hline
   		\cellcolor{black!5} \textbf{Prog-X}\phantom{[12]} & \cellcolor{black!5}\textbf{10.73} &  \cellcolor{black!5}\textbf{\phantom{1}8.73} & \cellcolor{black!5}14.38 & \cellcolor{black!5}\textbf{\phantom{1}6.86} & \cellcolor{black!5}\textbf{\phantom{1}5.91} & \cellcolor{black!5}\phantom{11}1.03 & \cellcolor{black!5}\textbf{\phantom{1}8.41} & \cellcolor{black!5}10.29 & \cellcolor{black!5}\textbf{0.02} & \cellcolor{black!5}\textbf{33.69} & \cellcolor{black!5}\phantom{1}\textbf{8.26} & \cellcolor{black!5}\phantom{1\;000}\textbf{9.36} \\
   		Multi-X\phantom{1}\cite{barath2018multix} & 17.13 & 12.23 & \phantom{1}\textbf{1.52} & \phantom{1}8.71 & \phantom{1}8.13 &  \phantom{11}\textbf{0.27} & 12.96 & 19.60 & 0.95 & 35.67 & 11.94 & \phantom{1}1\;407.39 \\ % \phantom{1}1.67 & 2.05 & 2.29 & \phantom{1}0.80 & 0.33
   		PEARL~\cite{isack2012energy} & 29.54 & 14.80 & \phantom{1}4.94 & 15.14 & \phantom{1}6.75 & \phantom{11}2.61 & 14.25 & 23.23 & 3.30 & 43.89 & 11.99 & \phantom{1}5\;142.39 \\ 
   		RPA~\cite{magri2015robust} & 17.11 & 11.08 & 10.24 & 23.54 & 13.42 & 622.87 & \phantom{1}9.16 & 11.26 & 4.92 & 55.47 & 11.71 & 10\;459.98 \\ 
   		RansaCov~\cite{magri2016multiple} & 55.61 & 12.42 & \phantom{1}2.33 & 66.88 & 18.44 & \phantom{1}17.69 & 11.13 & \phantom{1}\textbf{8.00} & 2.04 & 46.65 & 10.31 & \phantom{1}7\;914.00\\
   	    T-linkage~\cite{magri2014t} & 46.67 & 15.60 & \phantom{1}2.69 & 54.79 & 22.17 & \phantom{1}57.84 & 27.24 & 15.57 & 0.95 & 45.33 & 15.37 & \phantom{1}8\;423.31  \\
    \hline
    \hline      
\end{tabular}}
\caption{Average misclassification errors (in \%; 5 runs on each scene), their standard deviations and the processing times (in secs) for each problem: two-view motion fitting on the AdelaideRMF motion dataset (2nd--4th cols), homography estimation on the AdelaideRMF homography dataset (5th--7th), motion fitting on the Hopkins dataset (8th--10th), and simultaneous plane and cylinder fitting on the Multi-X dataset (11th--13th). Fixed parameters were used for all methods (see Table~\ref{tab:parameter_table} for Prog-X). Detailed results are in Fig.~\ref{fig:detailed_results}.}
\label{tab:summary_table}
\end{table*}

\subsection{Real world problems}

To evaluate the proposed method on real-world problems, we downloaded a number of publicly available datasets.
% for homography, two-view motion, and motion fitting. 
The error is the misclassification error (ME), i.e.\ the ratio of points assigned to the wrong cluster.
In Fig.~\ref{fig:real_results}, the first images are shown of image pairs which are some of the worst (left plots) and best (right) results of Prog-X running five times on each scene with a fixed parameters (reported in Table~\ref{tab:parameter_table}). 
The shown percentages are the misclassification errors. 
The white points are labeled as outliers, while the ones with color are assigned to an instance. 
It can be seen the error originates mostly from missing instances and, thus, the returned ones are usable. 

\noindent
\textbf{Two-view motion} fitting is evaluated on the AdelaideRMF motion dataset consisting of $21$ image pairs of different sizes and correspondences manually assigned to motion clusters. 
In this case, multiple fundamental matrices were fit. 
In the proposal step, the 7-point algorithm~\cite{hartley2003multiple} was applied for fitting to a minimal sample and the normalized 8-point method~\cite{hartley1997defense} for the polishing steps on non-minimal samples.
The average (of 5 runs with fixed parameters) errors and their standard deviations are shown in the first block of Table~\ref{tab:summary_table} for each method (from the $2$nd to the $4$th columns). 
Prog-X is superior to the competitors in all investigated properties.
Detailed results for each scene are shown in the top-left plot of Fig.~\ref{fig:detailed_results}. 
It can be seen that Prog-X is always the most or the second most accurate method.

\noindent
\textbf{Homography} fitting is evaluated on the AdelaideRMF homography dataset~\cite{wongiccv2011} used in most recent publications. AdelaideRMF consists of $19$ image pairs of different resolutions with ground truth point correspondences assigned manually to homographies. 
In the proposal step, the normalized 4-point algorithm~\cite{hartley2003multiple} was used for fitting to a minimal sample and also for the polishing steps. 
The results are shown in the second block of Table~\ref{tab:summary_table} (from the $5$th to the $7$th columns). 
Prog-X leads to the lowest average error and standard deviation. 
Similar results can be seen in the top-right plot of Fig.~\ref{fig:detailed_results} as before: Prog-X always leads to the lowest or second lowest misclassification errors.

\noindent
\textbf{Motion} segmentation is tested on the $155$ videos of the Hopkins dataset~\cite{tron2007benchmark}. 
The dataset consists of $155$ sequences divided into three categories: {\fontfamily{cmtt}\selectfont{checkerboard}}, {\fontfamily{cmtt}\selectfont{traffic}}, and {\fontfamily{cmtt}\selectfont{other}} sequences. 
The trajectories are inherently corrupted by noise, but no outliers are present.
Motion segmentation in videos is the retrieval of sets of points undergoing rigid motions contained in a dynamic scene captured by a moving camera.
It can be considered as a subspace segmentation under the assumption of affine cameras. 
For affine cameras, all feature trajectories associated with a single moving object lie in a 4D linear subspace in $\mathbb{R}^{2F}$, where $F$ is the number of frames~\cite{tron2007benchmark}.
The results are shown in the third block of Table~\ref{tab:summary_table} (from the $8$th to the $10$th columns). 
Prog-X leads to the lowest average errors. Its standard deviation is the second lowest.
Detailed results on the sequences of {\fontfamily{cmtt}\selectfont{traffic}} are put in the bottom of Fig.~\ref{fig:detailed_results}. 
%The error of Prog-X is approx.\ zero for most of the scenes. 

\noindent
\textbf{Plane and cylinder} fitting is evaluated on the dataset from~\cite{barath2018multix}.
It consists of LiDAR point clouds of traffic signs, their columns and the neighboring points. 
Points were manually assigned to signs (planes) and columns (cylinders). 
The proposal step of Prog-X alternately proposes cylinders and planes.
The results are in the last three columns of Table~\ref{tab:summary_table}. 
Prog-X obtains the most accurate results, but, most importantly, it is \textit{three orders of magnitude faster} than the second fastest method. 
The reason is the large number of points in the scenes (from $1\;260$ up to $52\;445$). 
The processing time of Prog-X, due to being dominated by a number of RANSAC runs, depends linearly on the point number. 
All of the other methods depend approximately quadratically on $n$.
In RPA, T-Linkage and RansaCov, the preference calculation has $\mathcal{O}(n^2)$ complexity. 
In Multi-X and PEARL, if at least $n$ initial instances are generated, the label space is already too large for the inner $\alpha$-expansion to finish early.

%Note that comparing the processing times of Matlab (RPA, T-Linkage, RansaCov) and C++ (PEARL, Multi-X, Prog-X) implementations might be unfair.
%However, it can be easily seen that the time of Prog-X is dominated by $k$ RANSAC runs and, thus, the implied complexity is roughly $\mathcal{O}(k m n)$, where $m$ is the number RANSAC iterations to a propose model and $n$ is the number of points. For PEARL and Multi-X, the dominant step is the $\alpha$-expansion, which is fairly slow in the case of large label space.
%Also, the complexity of RPA and T-Linkage is $\mathcal{O}(n^2)$ which is the reason of extreme running time for plane and cylinder fitting where each scene consists of tens of thousands of points. 

\section{Conclusion}

The Prog-X algorithm is proposed for geometric multi-class multi-model fitting. 
The method is fast and superior to the state-of-the-art in terms of accuracy in a synthetic environment and on publicly available real-world datasets for homography, two-motion, and motion segmentation. 
Additionally, it is an any-time algorithm. Therefore, whenever is interrupted, e.g.\ due to a strict time limit, the returned instances cover real and, likely, the most dominant ones.
The termination criterion, adopted from RANSAC, makes Prog-X robust to the inlier-outlier ratio.
%
%The source code will be made publicly available after publication. 

{\small
\bibliographystyle{ieee}
\bibliography{egbib}

\begin{thebibliography}{10}\itemsep=-1pt

\bibitem{amayo2018geometric}
P.~Amayo, P.~Pini{\'e}s, L.~M. Paz, and P.~Newman.
\newblock Geometric multi-model fitting with a convex relaxation algorithm.
\newblock In {\em Proceedings of the IEEE Conference on Computer Vision and
  Pattern Recognition}, pages 8138--8146, 2018.

\bibitem{barath2018graph}
D.~Bar\'ath and J.~Matas.
\newblock Graph-cut ransac.
\newblock {\em IEEE Conference on Computer Vision and Pattern Recognition},
  2018.

\bibitem{barath2018multix}
D.~Barath and J.~Matas.
\newblock Multi-class model fitting by energy minimization and mode-seeking.
\newblock In {\em European Conference on Computer Vision}, 2018.

\bibitem{boykov2004experimental}
Y.~Boykov and V.~Kolmogorov.
\newblock An experimental comparison of min-cut/max-flow algorithms for energy
  minimization in vision.
\newblock {\em Pattern Analysis and Machine Intelligence}, 2004.

\bibitem{boykov2001fast}
Y.~Boykov, O.~Veksler, and R.~Zabih.
\newblock Fast approximate energy minimization via graph cuts.
\newblock {\em Pattern Analysis and Machine Intelligence}, 2001.

\bibitem{broder1997resemblance}
A.~Z. Broder.
\newblock On the resemblance and containment of documents.
\newblock In {\em Compression and complexity of sequences 1997. proceedings},
  pages 21--29. IEEE, 1997.

\bibitem{delong2012minimizing}
A.~Delong, L.~Gorelick, O.~Veksler, and Y.~Boykov.
\newblock Minimizing energies with hierarchical costs.
\newblock {\em International Journal of Computer Vision}, 2012.

\bibitem{delong2012fast}
A.~Delong, A.~Osokin, H.~N. Isack, and Y.~Boykov.
\newblock Fast approximate energy minimization with label costs.
\newblock {\em International journal of computer vision}, 2012.

\bibitem{fischler1981random}
M.~A. Fischler and R.~C. Bolles.
\newblock Random sample consensus: a paradigm for model fitting with
  applications to image analysis and automated cartography.
\newblock {\em Communications of the ACM}, 1981.

\bibitem{guil1997lower}
N.~Guil and E.~L. Zapata.
\newblock Lower order circle and ellipse hough transform.
\newblock {\em Pattern Recognition}, 1997.

\bibitem{hartley2003multiple}
R.~Hartley and A.~Zisserman.
\newblock {\em Multiple view geometry in computer vision}.
\newblock 2003.

\bibitem{hartley1997defense}
R.~I. Hartley.
\newblock In defense of the eight-point algorithm.
\newblock {\em Transactions on Pattern Analysis and Machine Intelligence},
  1997.

\bibitem{vc1962method}
P.~V.~C. Hough.
\newblock Method and means for recognizing complex patterns, 1962.

\bibitem{illingworth1988survey}
J.~Illingworth and J.~Kittler.
\newblock A survey of the hough transform.
\newblock {\em Computer Vision, Graphics, and Image Processing}, 1988.

\bibitem{isack2012energy}
H.~Isack and Y.~Boykov.
\newblock Energy-based geometric multi-model fitting.
\newblock {\em International Journal on Computer Vision}, 2012.

\bibitem{kanazawa2004detection}
Y.~Kanazawa and H.~Kawakami.
\newblock Detection of planar regions with uncalibrated stereo using
  distributions of feature points.
\newblock In {\em British Machine Vision Conference}, 2004.

\bibitem{magri2014t}
L.~Magri and A.~Fusiello.
\newblock {T-Linkage}: A continuous relaxation of {J-Linkage} for multi-model
  fitting.
\newblock In {\em Conference on Computer Vision and Pattern Recognition}, 2014.

\bibitem{magri2015robust}
L.~Magri and A.~Fusiello.
\newblock Robust multiple model fitting with preference analysis and low-rank
  approximation.
\newblock In {\em British Machine Vision Conference}, 2015.

\bibitem{magri2016multiple}
L.~Magri and A.~Fusiello.
\newblock Multiple model fitting as a set coverage problem.
\newblock In {\em Conference on Computer Vision and Pattern Recognition}, 2016.

\bibitem{matas2000robust}
J.~Matas, C.~Galambos, and J.~Kittler.
\newblock Robust detection of lines using the progressive probabilistic hough
  transform.
\newblock {\em Computer Vision and Image Understanding}, 2000.

\bibitem{pham2014interacting}
T.~T. Pham, T.-J. Chin, K.~Schindler, and D.~Suter.
\newblock Interacting geometric priors for robust multi-model fitting.
\newblock {\em {TIP}}, 2014.

\bibitem{nasuto2002napsac}
D.~R.~Myatt, P.~Torr, S.~Nasuto, J.~Bishop, and R.~Craddock.
\newblock {NAPSAC}: High noise, high dimensional robust estimation - it’s in
  the bag.
\newblock 2002.

\bibitem{rosin1993ellipse}
P.~L. Rosin.
\newblock Ellipse fitting by accumulating five-point fits.
\newblock {\em Pattern Recognition Letters}, 1993.

\bibitem{tanimoto1958elementary}
T.~T. Tanimoto.
\newblock Elementary mathematical theory of classification and prediction.
\newblock 1958.

\bibitem{toldo2008robust}
R.~Toldo and A.~Fusiello.
\newblock Robust multiple structures estimation with {J-Linkage}.
\newblock In {\em European Conference on Computer Vision}, 2008.

\bibitem{torr2002bayesian}
P.~H.~S. Torr.
\newblock Bayesian model estimation and selection for epipolar geometry and
  generic manifold fitting.
\newblock {\em International Journal of Computer Vision}, 50(1):35--61, 2002.

\bibitem{tron2007benchmark}
R.~Tron and R.~Vidal.
\newblock A benchmark for the comparison of 3-d motion segmentation algorithms.
\newblock In {\em Conference on Computer Vision and Pattern Recognition}, 2007.

\bibitem{vincent2001detecting}
E.~Vincent and R.~Lagani{\'e}re.
\newblock Detecting planar homographies in an image pair.
\newblock In {\em International Symposium on Image and Signal Processing and
  Analysis}, 2001.

\bibitem{wang2015mode}
H.~Wang, G.~Xiao, Y.~Yan, and D.~Suter.
\newblock Mode-seeking on hypergraphs for robust geometric model fitting.
\newblock In {\em International Conference of Computer Vision}, 2015.

\bibitem{wang2018searching}
h.~Wang, G.~Xiao, Y.~Yan, and D.~Suter.
\newblock Searching for representative modes on hypergraphs for robust
  geometric model fitting.
\newblock {\em Transactions on Pattern Analysis and Machine Intelligence},
  2018.

\bibitem{wongiccv2011}
H.~S. Wong, T.-J. Chin, J.~Yu, and D.~Suter.
\newblock Dynamic and hierarchical multi-structure geometric model fitting.
\newblock In {\em International Conference on Computer Vision}, 2011.

\bibitem{xu1990new}
L.~Xu, E.~Oja, and P.~Kultanen.
\newblock A new curve detection method: randomized hough transform (rht).
\newblock {\em Pattern Recognition Letters}, 1990.

\bibitem{zhang2007nonparametric}
W.~Zhang and J.~Koseck{\'a}.
\newblock Nonparametric estimation of multiple structures with outliers.
\newblock In {\em Dynamical Vision}. 2007.

\bibitem{zuliani2005multiransac}
M.~Zuliani, C.~S. Kenney, and B.~Manjunath.
\newblock The multiransac algorithm and its application to detect planar
  homographies.
\newblock In {\em {ICIP}}. {IEEE}, 2005.

\end{thebibliography}
}

\end{document}